\documentclass[sigconf]{acmart}
\AtBeginDocument{%
  }

\setcopyright{acmlicensed}

\copyrightyear{2026}
\acmYear{2026}
\setcopyright{cc}
\setcctype{by}
\acmConference[MM '26]{Proceedings of the 34th ACM International Conference on Multimedia}{November 10--14, 2026}{Rio de Janeiro, Brazil}
\acmBooktitle{Proceedings of the 34th ACM International Conference on Multimedia (MM '26), November 10--14, 2026, Rio de Janeiro, Brazil}
\acmDOI{10.1145/3767308.3836344}
\acmISBN{979-8-4007-2213-4/2026/11}

\usepackage{booktabs}
\usepackage{multirow}
\usepackage{graphicx}
\usepackage{colortbl}
\usepackage{makecell}
\usepackage{tikz}
\usepackage{pifont}
\usepackage{siunitx}
\usepackage{arydshln}
\usepackage{balance} 

\definecolor{bestrow}{RGB}{248, 251, 227}
\definecolor{checkgreen}{RGB}{34, 139, 34}
\definecolor{crossred}{RGB}{200, 40, 40}
\definecolor{deepgreen}{RGB}{0,120,80}
\definecolor{deepred}{RGB}{180,30,30}
\definecolor{upcolor}{HTML}{E55C7A}
\definecolor{downcolor}{HTML}{4A90E2}
\definecolor{c1}{HTML}{9374D2}
\definecolor{c2}{HTML}{BE74C1}
\definecolor{c3}{HTML}{DF74A4}
\definecolor{c4}{HTML}{EF748A}
\definecolor{rowbg}{HTML}{F5F9D6}

\newcommand{\cmark}{\textcolor{deepgreen}{\ensuremath{\checkmark}}}
\newcommand{\xmark}{\textcolor{deepred}{\ensuremath{\times}}}

\usepackage[ruled,vlined,linesnumbered]{algorithm2e}
\usepackage{xcolor}
\usepackage{graphicx}

\newlength{\AlgoRowWidth}
\newcommand{\AlgoRow}[2]{%
  \begingroup
  \setlength{\AlgoRowWidth}{\dimexpr\hsize-\leftskip-0.8em\relax}%
  \setlength{\fboxsep}{0pt}%
  \colorbox{#1}{\makebox[\AlgoRowWidth][l]{\strut #2\hfill}}%
  \endgroup
}

\begin{document}

\title{Local Epistemic Uncertainty Guided Active Sampling for Plug-and-play Diffusive Image Restoration}

\author{Jiaqi Zhang}
\email{3230602065@stmail.ujs.edu.cn}
\orcid{0009-0002-0924-9702}
\affiliation{%
  \institution{
  % Department of Computer Science\\
  Jiangsu University}
  \city{Zhenjiang}
  \country{China}
}

\author{Zheng Pang}
\email{3230613024@stmail.ujs.edu.cn}
\orcid{0009-0003-6606-3151}
\affiliation{%
  \institution{
  % Department of Computer Science\\
  Jiangsu University}
  \city{Zhenjiang}
  \country{China}
}

\author{Rongrong Gao}
\email{3230608056@stmail.ujs.edu.cn}
\orcid{0009-0009-6376-3623}
\affiliation{%
  \institution{
  % Department of Computer Science\\
  Jiangsu University}
  \city{Zhenjiang}
  \country{China}
}

\author{Qiyuan Zhang}
\email{3230342037@stu.xaut.edu.cn}
\orcid{0009-0003-9422-4688}
\affiliation{%
  \institution{
  % Packaging and Digital Media College\\
  Xi'an University of Technology}
  \city{Xi'an}
  \country{China}
}

\author{Yang Yang}
\authornote{Yang Yang is the Corresponding Author.}
\email{yyoung@ujs.edu.cn}
\orcid{0000-0001-8782-4819}
\affiliation{%
  \institution{
  % Department of Computer Science\\
  Jiangsu University}
  \city{Zhenjiang}
  \country{China}
}

\begin{abstract}
Diffusion models have demonstrated remarkable effectiveness in image restoration tasks. However, when guiding image reconstruction, existing Diffusion Model-based Image Restoration (DMIR) methods typically rely on fixed data constraints and uniform step sizes, thereby overlooking the dynamic nature of the generative process. Such rigid designs render the models vulnerable to spatially non-uniform degradations, thus resulting in structural distortions and loss of fine details. Meanwhile, uniform step sizes introduce computational redundancy, whereas naïve step reduction strategies tend to accumulate approximation errors. To address these limitations, we propose a Local Epistemic Uncertainty Guided Active Sampling framework (LEADer). In the spatial domain, LEADer leverages pixel-wise uncertainty to dynamically modulate the prior strength within the null space, which effectively balances detail preservation and artifact suppression. In the temporal domain, it quantifies sampling stability via the uncertainty trace to enable adaptive trajectory pruning, thereby accelerating convergence. Theoretical proofs demonstrate that our framework achieves strict data consistency, while the trajectory pruning strategy admits a deterministic error bound, thereby guaranteeing stable convergence under skip sampling. Notably, our plug-and-play method can be seamlessly integrated into various DMIR baselines. Extensive experiments show that LEADer improves the performance of multiple state-of-the-art DMIR methods, while significantly reducing sampling time with negligible memory overhead. 
Code is available at \href{https://www.jiaqizhang-sengoku.com/Code/LEADer}{here}.
\end{abstract}




\maketitle
\section{Introduction}
Image restoration~\cite{DBLP:journals/tip/DabovFKE07, DBLP:conf/cvpr/GuZZF14, DBLP:journals/pami/DongLHT16, DBLP:conf/cvpr/LedigTHCCAATTWS17, DBLP:conf/iccvw/LiangCSZGT21, DPG} aims to reconstruct clear, natural images from degraded observations corrupted by noise, blur, or low resolution, serving as a fundamental problem in computer vision and medical imaging~\cite{DBLP:journals/pami/Wang0H21, DBLP:journals/tmi/WangYMF18, DBLP:journals/spm/McCannJU17}. Recently, deep learning-based methods have achieved remarkable performance on various specific image restoration tasks~\cite{DBLP:journals/tip/ZhangZCM017, DBLP:conf/iccvw/LiangCSZGT21, DBLP:conf/cvpr/ZamirA0HK022, DBLP:conf/eccv/ChenCZS22, DBLP:journals/pami/DongLHT16}. However, these methods, optimized for specific degradation patterns, typically rely on massive paired training data. Moreover, this task-specific paradigm often suffers from limited generalization capabilities~\cite{DBLP:journals/pami/ZhangLZZGT22}.

Recently, diffusion models~\cite{DBLP:conf/icml/Sohl-DicksteinW15, DBLP:conf/nips/HoJA20, DBLP:conf/nips/SongE19, DBLP:conf/iclr/0011SKKEP21, DBLP:journals/pami/CroitoruHIS23, DBLP:journals/csur/YangZSHXZZCY24, DG-SCL} have demonstrated powerful capabilities in modeling complex data distributions and have been successfully applied to diverse image generation tasks~\cite{DBLP:conf/cvpr/RombachBLEO22, DBLP:conf/nips/DhariwalN21, AgentTraces}. Consequently, researchers have begun exploring pre-trained unconditional diffusion models as universal priors to solve image restoration inverse problems~\cite{DBLP:conf/cvpr/FeiLPZYLZ023}. Diffusion Model-based Image Restoration (DMIR) methods typically adopt a posterior sampling paradigm: they directly utilize pre-trained diffusion models to fit the prior distribution of clear natural images, incorporating likelihood guidance~\cite{DBLP:conf/iclr/0011S0E22, DBLP:conf/nips/ChungSRY22, DBLP:conf/iclr/ChungKMKY23, DBLP:conf/mm/WuHZ0LLZ0024, DBLP:conf/mm/WuZHZ025, DBLP:conf/icml/AlkhouriLHD0RW25, DBLP:conf/iclr/MardaniSKV24, DBLP:conf/cvpr/0011GF00KW25} or data consistency projection~\cite{DBLP:conf/cvpr/LugmayrDRYTG22, DBLP:conf/nips/KawarEES22, DBLP:conf/cvpr/ZhuZLCWTG23, DBLP:conf/iclr/WangYZ23, DBLP:conf/nips/ZhangZJLG24, DBLP:conf/cvpr/GarberT24, DBLP:journals/kbs/YangZZZ26, DBLP:journals/corr/abs-2511-09965, DBLP:conf/nips/ChungKY23, DBLP:journals/corr/abs-2411-15295} during the reverse sampling process. This enables a single pre-trained model to adapt to diverse degradation scenarios, which effectively enhances the generalization capability of restoration methods.

However, during reverse sampling, existing DMIR methods typically employ fixed data constraint weights and uniform time steps. This setting overlooks the dynamic nature of image generation and introduce two main challenges. Spatially, since local degradations in real images are often uneven, globally fixed weights struggle to accommodate varying regional restoration needs, thus leading to compromised details or inaccurate local structures. Temporally, fixed time steps cause additional computational cost during the stationary phase of sampling, whereas directly reducing steps introduces additional discretization errors that compromise overall restoration quality~\cite{DBLP:conf/nips/KarrasAAL22, DBLP:conf/nips/0011ZB0L022}.

\begin{figure}[!t]
\centering
\includegraphics[width=0.45\textwidth]{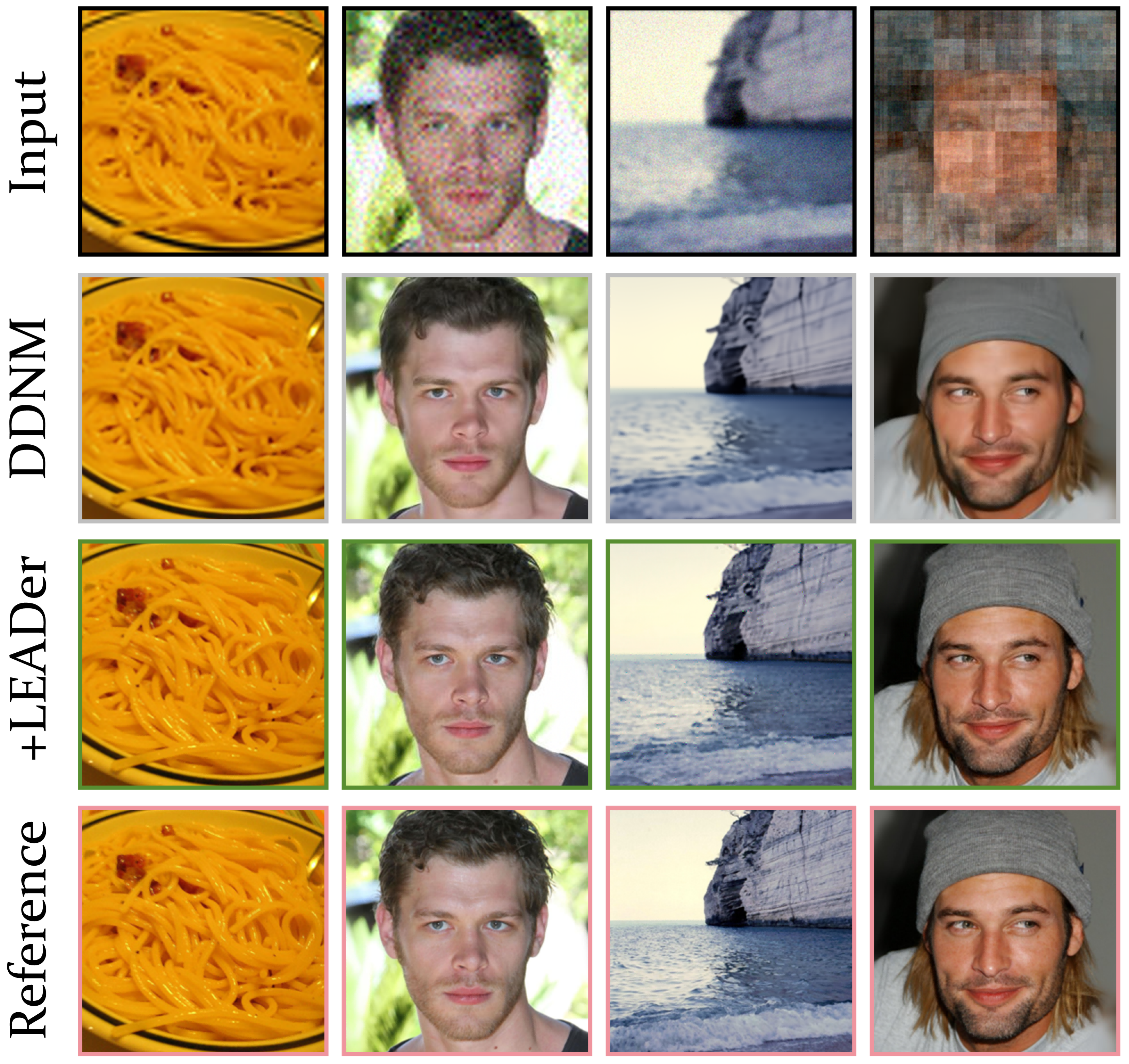}
\caption{Qualitative comparisons of the proposed method with plug-and-play adaptation across different tasks.}
\vspace{-5mm}
\label{Figure1}
\end{figure}

In response to these challenges, we propose an active diffusion sampling framework guided by local epistemic uncertainty, termed LEADer. Specifically, to ensure structural consistency, we propose Uncertainty-Calibrated Prior Modulation (UCPM), which leverages pixel-level uncertainty to dynamically regulate the strength of prior within the null space, thereby adaptively balancing detail preservation and artifact suppression. Furthermore, to improve inference efficiency, we propose State-Aware Trajectory Pruning (SATP). This technique evaluates the stability of the sampling process via the trace of uncertainty and executes step-skipping within an controlled error tolerance, which effectively accelerates convergence and improves quality, as illustrated in Figure~\ref{Figure1}. In addition, theoretical analysis demonstrates that our method guarantees data fidelity while providing a controlled error bound for the skip sampling process. The main contributions of this paper are as follows:
\begin{itemize}
    \item We propose a novel perspective on the diffusion reverse process based on local epistemic uncertainty, which reveals the spatiotemporal limitations of the  sampling strategies in existing DMIR methods.
    \item We propose an active diffusion sampling framework that proactively resolves local distortion and sampling redundancy via uncertainty-calibrated prior modulation and state-aware trajectory pruning.
    \item Extensive experimental results demonstrate that our proposed plug-and-play framework effectively enhances image restoration quality and performance while improving inference efficiency.
\end{itemize}

\section{Background}
DMIR methods have emerged as the mainstream for zero-shot solvers in image inverse problems. The core challenge lies in injecting specific measurement constraints into the reverse denoising process. Depending on the intervention mechanisms, existing methods are primarily categorised into the following two classes:

\noindent\textbf{Gradient-based Optimization Guidance:} Rooted in Bayesian posterior sampling theory~\cite{DBLP:conf/iclr/ChungKMKY23, DBLP:conf/iclr/0011S0E22, DBLP:conf/iclr/MardaniSKV24}, these methods approximate the measurement likelihood gradient using the clean image estimate at the current time step to guide reverse sampling. Notably, DPS~\cite{DBLP:conf/iclr/ChungKMKY23} directly enforces data consistency using the gradient of the $l_2$-norm of the measurement error. Subsequent works have focused on improving guidance accuracy and stability. For example, MCG~\cite{DBLP:conf/nips/ChungSRY22} incorporates manifold projection constraints, while $\Pi$GDM~\cite{DBLP:conf/iclr/SongVMK23} enhances robustness via pseudo-inverse matrices and Jacobian computations. ZAPS~\cite{DBLP:conf/eccv/AlcalarA24} further proposes an alternative guidance scheme to improve performance over DPS~\cite{DBLP:conf/iclr/ChungKMKY23}. More recent studies investigate the numerical instability and bias arising during the guidance process. For instance, SITCOM~\cite{DBLP:conf/icml/AlkhouriLHD0RW25} and SPGD~\cite{DBLP:conf/mm/WuZHZ025} introduce explicit gradient control to stabilize optimization. In parallel, DCDP~\cite{li2024decoupled} and DAPS~\cite{DBLP:conf/cvpr/ZhangCBMA025} refine the sampling trajectory by decoupling the purification process to mitigate prior bias, while DPPS~\cite{DBLP:conf/mm/WuHZ0LLZ0024} employs proximal operators to handle non-smooth constraints, thereby improving flexibility under complex degradations.

\noindent\textbf{Null-Space Orthogonal Projection:} In contrast to gradient-based guidance, these methods decompose the image space into range and null spaces via orthogonal projection, and iteratively refine the null-space components during reverse diffusion. Representative approaches such as DDNM~\cite{DBLP:conf/iclr/WangYZ23} and DDRM~\cite{DBLP:conf/nips/KawarEES22} achieve strict data consistency through pseudo-inverse projection and singular value decomposition (SVD), respectively. Building upon this foundation, subsequent works extend the framework to improve flexibility and guidance quality. DiffPIR~\cite{DBLP:conf/cvpr/ZhuZLCWTG23} integrates Half-Quadratic Splitting (HQS) to reformulate measurement constraints as proximal mappings, while DDPG~\cite{DBLP:conf/cvpr/GarberT24} introduces a preconditioned guidance trajectory that evolves with sampling steps, which effectively bridges back-projection and least-squares gradient formulations. More recent studies focus on enhancing global consistency and structural regularity. For example, ProjDiff~\cite{DBLP:conf/nips/ZhangZJLG24} constructs a bivariate constrained optimization framework to better exploit denoising priors, EquS~\cite{DBLP:journals/corr/abs-2511-09965} utilizes equivariant sampling to constrain manifold regularity, and PIRP~\cite{DBLP:journals/kbs/YangZZZ26} incorporates parameterized gradient priors to elevate overall perception and fidelity.

\begin{figure*}[!t]
\centering
\includegraphics[width=0.98\textwidth]{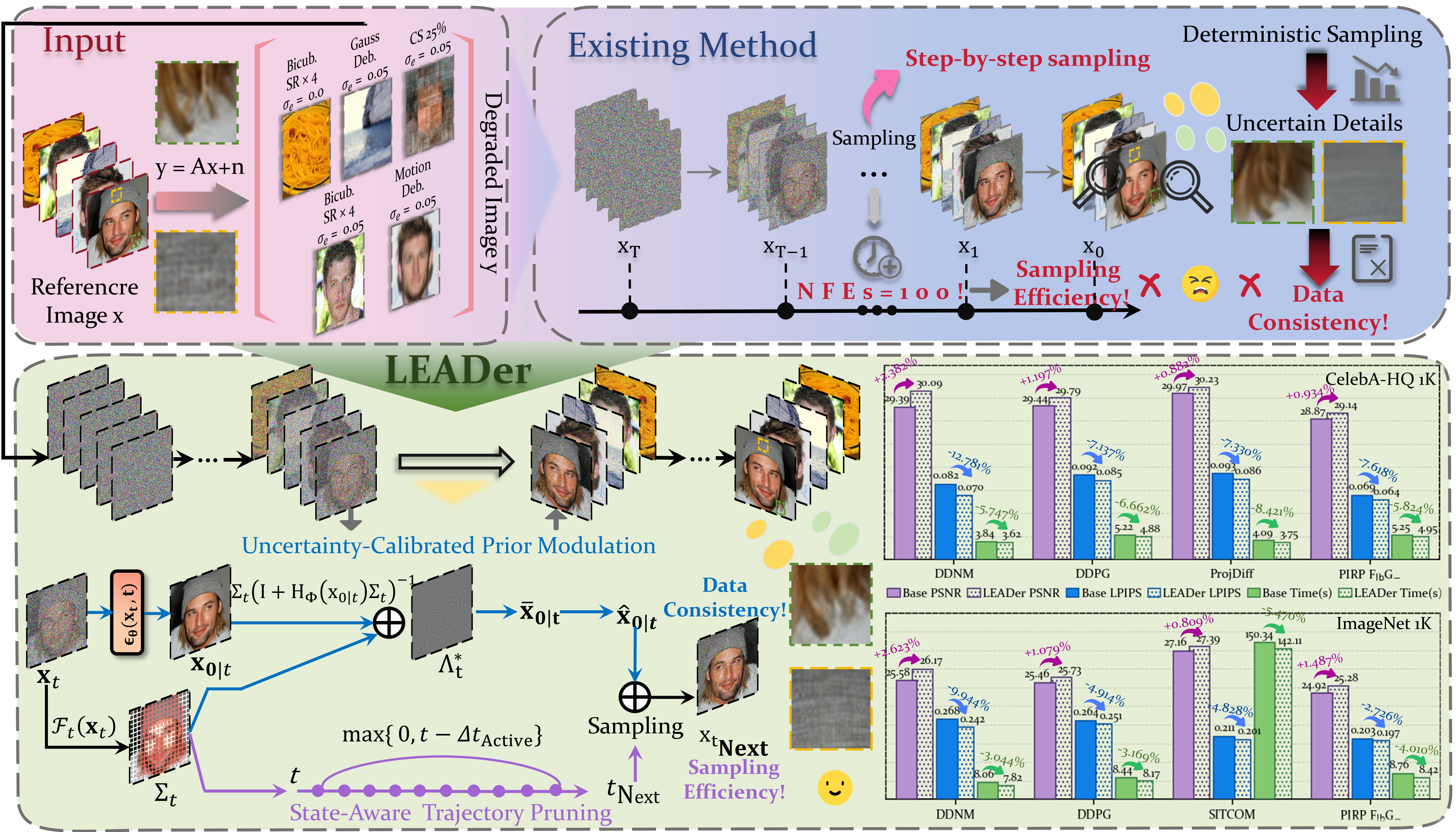} 
\caption{{Architecture of LEADer.} Existing DMIR methods rely on fixed step-by-step sampling with uniform constraints, leading to computational redundancy and detail distortion in uncertain regions. To address this, LEADer quantifies local epistemic uncertainty $\boldsymbol{\Sigma}_t$ to guide reverse sampling. Spatially, UCPM utilizes $\boldsymbol{\Sigma}_t$ to adaptively modulate prior constraints for strict data consistency. Temporally, SATP computes an adaptive step size $\Delta t_\text{Active}$ based on the uncertainty trace to skip redundant iterations. As a result, LEADer improves the restoration quality of various baselines while reducing sampling time.}
\label{Figure2}
\vspace{-5mm}
\end{figure*}

\section{Preliminary}
\label{Section3}
\textbf{Problem Formulation:} The image restoration task aims to recover a target clean image $\mathbf{x}_0 \in \mathbb{R}^N$ from a degraded observation $\mathbf{y} \in \mathbb{R}^M$. This physical degradation process is typically modeled as a linear inverse problem:
\begin{equation}
    \mathbf{y} = \mathbf{A}\mathbf{x} + \mathbf{n},
\end{equation}
where $\mathbf{y} \in \mathbb{R}^M$ is the degraded image, and $\mathbf{x}_0 \in \mathbb{R}^N$ represents the high-quality image to be recovered. $\mathbf{A} \in \mathbb{R}^{M \times N}$ is a known linear operator, such as a bicubic downsampler in image super-resolution. Finally, $\mathbf{n} \sim \mathcal{N}(\mathbf{0}, \sigma_y^2 \mathbf{I})$ is additive Gaussian noise~\cite{DBLP:conf/nips/KawarEES22}, which represents uncertainty or disturbance in the observation $\mathbf{y}$.

Within the zero-shot diffusion framework, restoration is achieved without retraining on paired degraded data. Instead, a pre-trained unconditional diffusion model serves as a generative prior. At any time step $t \in (0, T]$ during reverse sampling, given the current noisy variable $\mathbf{x}_t$, the diffusion model predicts the noise component via a denoising neural network $\boldsymbol{\epsilon}_\theta(\mathbf{x}_t, t)$. Based on Tweedie's formula~\cite{stein1981estimation, efron2011tweedie}, the corresponding estimate of the clean image is:
\begin{equation}
    \mathbf{x}_{0|t} = \frac{\mathbf{x}_t - \sqrt{1 - \bar{\alpha}_t} \boldsymbol{\epsilon}_\theta(\mathbf{x}_t, t)}{\sqrt{\bar{\alpha}_t}},
\label{x0t}
\end{equation}
where $\bar{\alpha}_t$ is a predefined noise schedule parameter. 

To enforce data consistency with the physical observation, the Denoising Diffusion Null-Space Model~\cite{DBLP:conf/iclr/WangYZ23} introduces an orthogonal decomposition of the image space into range and null spaces:
\begin{equation}
    \hat{\mathbf{x}}_{0|t} = \mathbf{A}^{\dagger}\mathbf{y} + (\mathbf{I} - \mathbf{A}^{\dagger}\mathbf{A})\mathbf{x}_{0|t},
\end{equation}
where $\mathbf{A}^\dagger$ is the pseudo-inverse of the degradation matrix. $\mathbf{A}^{\dagger}\mathbf{y}$ enforces data fidelity in the range space, while $(\mathbf{I} - \mathbf{A}^\dagger \mathbf{A})\mathbf{x}_{0|t}$ projects $\mathbf{x}_{0|t}$ onto the null space for prior-guided refinement.

\noindent\textbf{Diffusion Models:} Denoising Diffusion Probabilistic Models (DDPM) \cite{DBLP:conf/nips/HoJA20} learn data generation by reversing a diffusion process that progressively corrupts data through noise addition. This framework consists of a forward process and a reverse process.

The forward process is a fixed Markov chain that gradually perturbs the data over $T$ time steps. Let $\beta_t$ denote the noise schedule. At each step $t$, the noisy state $\mathbf{x}_t$ is given by:
\begin{equation}
    \mathbf{x}_t = \sqrt{1 - \beta_t}\mathbf{x}_{t-1} + \sqrt{\beta_t}{\boldsymbol{\epsilon}}_{t-1},
\end{equation}
where ${\boldsymbol{\epsilon}}_{t-1} \sim \mathcal{N}(\mathbf{0}, \mathbf{I})$. Let the cumulative noise up to time step $t$ be $\bar{\alpha}_t = \prod_{i=1}^t \alpha_i$, then we have:
\begin{equation}
    \mathbf{x}_t = \sqrt{\bar{\alpha}_t}\mathbf{x}_0 + \sqrt{1 - \bar{\alpha}_t}{\boldsymbol{\epsilon}}.
\end{equation}

The reverse process aims to reconstruct the original clean image $\mathbf{x}_0$ from the noisy state $\mathbf{x}_t$. Mathematically, this is described by:
\begin{equation}
    \mathbf{x}_{t-1} = \frac{1}{\sqrt{\alpha_t}} \left( \mathbf{x}_t - \frac{1 - \alpha_t}{\sqrt{1 - \bar{\alpha}_t}} {\boldsymbol{\epsilon}}_\theta(\mathbf{x}_t, t) \right) + \sigma_t {\boldsymbol{\epsilon}}_t,
\end{equation}
where $\boldsymbol{\epsilon}_\theta(\mathbf{x}_t, t)$ denotes the noise predicted by a neural network parameterized by $\theta$, and $\boldsymbol{\epsilon}_t \sim \mathcal{N}(\mathbf{0}, \mathbf{I})$ is the Gaussian noise.

Despite their strong generative capability, DDPM suffers from high computational cost due to the large number of sampling steps. To address this limitation, \cite{DBLP:conf/iclr/SongME21} introduces Denoising Diffusion Implicit Models (DDIM), which relax the Markovian assumption, and enable more efficient sampling with fewer steps. Based on the estimate $\mathbf{x}_{0|t}$ in Eq.~\eqref{x0t}, the DDIM update rule is:
\begin{equation}
    \mathbf{x}_{t-1} = \sqrt{\bar{\alpha}_{t-1}}\mathbf{x}_{0|t} + \sqrt{1 - \bar{\alpha}_{t-1} - \sigma_t^2} {\boldsymbol{\epsilon}}_\theta(\mathbf{x}_t, t) + \sigma_t {\boldsymbol{\epsilon}}_t.
\end{equation}

\section{Proposed Method}
\subsection{Motivation}
Existing zero-shot solvers typically employ globally fixed prior weights and uniform time steps when intervening in the reverse diffusion process. Such a rigid design implicitly treats sampling as a static procedure and overlooks the inherently dynamic characteristics of diffusion models across different spatial regions and temporal evolution stages. This gives rise to two key issues:

\noindent\textbf{Perception-Distortion Trade-off:} The level of degradation and the difficulty of restoration varies significantly across different image regions~\cite{DBLP:conf/cvpr/BlauM18, he2026cinematte, zhang2026drive}. Applying globally uniform constraint weights fails to account for this heterogeneity. Specifically, in regions where fine details can be already recovered, overly strong priors tend to over-smooth structures. On the other hand, in severely degraded areas where prior-driven inference is less reliable, insufficient constraints can lead to deviation from the true underlying structures~\cite{DBLP:conf/icml/NingSPCC23, he2023fast, zhang2026if}.

\noindent\textbf{Sampling Redundancy:} From the perspective of ordinary differential equation (ODE), the reverse diffusion trajectory evolves with non-uniform dynamics. However, existing methods typically adopt fixed, uniform step sizes. This static temporal scheduling ignores the varying demands of different sampling stages. Once the global structure is established, continuing dense step-by-step updates introduces substantial computational redundancy; whereas naively increasing step sizes can accumulate errors during detail reconstruction, which ultimately degrades the  quality~\cite{DBLP:conf/nips/KarrasAAL22, zheng2022leveraging, tao2024cntools, 11622841}.

To address these issues, we argue that the diffusion sampling steps should be state-aware. Specifically, we quantify local epistemic uncertainty at each spatial location and sampling step, then we use it as a unified modulation signal. In the spatial dimension, this uncertainty facilitates pixel-level on-demand regularization modulation to balance detail preservation and artifact suppression (see Section~\ref{Section4.3}). In the temporal dimension, it serves as an error control metric to guide adaptive step sizing (see Section~\ref{Section4.4}). The overall architecture of our method is illustrated in Figure~\ref{Figure2}.

\subsection{Local Epistemic Uncertainty Quantification}
\label{Section4.2}
At any time step $t$ in the reverse dynamics, the noisy observation $\mathbf{x}_t$ embodies two fundamentally different types of uncertainty: the inherent aleatoric uncertainty introduced by the forward diffusion process, and the epistemic uncertainty arising from missing information, particularly within the null space. In this paper, we focus on quantifying the latter from an information geometry perspective.

Based on the marginal distribution $q(\mathbf{x}_t|\mathbf{x}_0)$ of the forward noise-adding process, the Observed Fisher Information Matrix (FIM) $\mathcal{F}_t(\mathbf{x}_t)\in \mathbb{R}^{N \times N}$ of the pre-trained diffusion model at the current state $\mathbf{x}_t$ is defined as:
\begin{equation}
    \mathcal{F}_t(\mathbf{x}_t) = - \nabla_{\mathbf{x}_t}^2 \log p_t(\mathbf{x}_t),
\end{equation}
where $\mathcal{F}_t(\mathbf{x}_t)$ represents the model's information sensitivity and structural certainty regarding the local features at this state. A higher value indicates that the model has stronger prior confidence in high-frequency details, whereas a lower value implies higher generative ambiguity in that region. As detailed in Section~\ref{Section3}, the posterior expectation estimate $\mathbf{x}_{0|t}$ of the target clean image $\mathbf{x}_0$ is:
\begin{equation}
    \mathbf{x}_{0|t} = \mathbb{E}[\mathbf{x}_0 | \mathbf{x}_t] = \frac{1}{\sqrt{\bar{\alpha}_t}} \left( \mathbf{x}_t + (1-\bar{\alpha}_t) \nabla_{\mathbf{x}_t} \log p_t(\mathbf{x}_t) \right).
\end{equation}

To quantify the prediction variance of the model given $\mathbf{x}_t$, we perform a second-order Taylor expansion on the posterior distribution to derive the analytical relationship between the posterior covariance matrix $\boldsymbol{\Sigma}_t$ and FIM:
\begin{equation}
    \boldsymbol{\Sigma}_t = \mathrm{Cov}(\mathbf{x}_0 | \mathbf{x}_t) = \frac{1-\bar{\alpha}_t}{\bar{\alpha}_t} \left[ \mathbf{I} - (1-\bar{\alpha}_t) \mathcal{F}_t(\mathbf{x}_t) \right].
\end{equation}
This decomposition provides a clear interpretation. The leading term $\frac{1-\bar{\alpha}_t}{\bar{\alpha}_t} \mathbf{I}$ represents the inherent upper bound of the aleatoric uncertainty at time step $t$, while the second term $(1-\bar{\alpha}_t) \mathcal{F}_t(\mathbf{x}_t)$ reflects the Fisher information gain provided by the pre-trained model via the score function. Consequently, the diagonal matrix $\boldsymbol{\Sigma}_t$ serve as an effective measure of local epistemic uncertainty. Specifically, when $\mathcal{F}_t(\mathbf{x}_t)$ is large, the model exhibits high confidence in the reconstructed structures, thus leading to an epistemic uncertainty of $\boldsymbol{\Sigma}_t \to \mathbf{0}$. Conversely, in regions with limited observational constraints, particularly within the null-space, the Fisher information diminishes, thus resulting in a higher $\boldsymbol{\Sigma}_t$.

\subsection{Uncertainty-Calibrated Prior Modulation}
\label{Section4.3}
To address the limitations of globally uniform weighting, we propose Uncertainty-Calibrated Prior Modulation (UCPM). By leveraging the extracted local epistemic uncertainty $\boldsymbol{\Sigma}_t$ as a spatial guidance signal, UCPM dynamically modulates the strength of physical priors $\Phi(\mathbf{x})$ (e.g., gradient and structural priors) at the pixel level. Specifically, we cast the problem as a Maximum A Posteriori (MAP) optimization objective within the local state space~\cite{lan2025mappo}:
\begin{equation}
    \mathcal{J}(\mathbf{x}) = \frac{1}{2} (\mathbf{x} - \mathbf{x}_{0|t})^T \boldsymbol{\Sigma}_t^{-1} (\mathbf{x} - \mathbf{x}_{0|t}) + \Phi(\mathbf{x}).
\end{equation}
Setting the first-order optimality condition $\nabla \mathcal{J}(\mathbf{x}) = \mathbf{0}$ yields:
\begin{equation}
    \boldsymbol{\Sigma}_t^{-1} (\mathbf{x} - \mathbf{x}_{0|t}) + \nabla \Phi(\mathbf{x}) = \mathbf{0}.
\end{equation}
To obtain a tractable solution, we linearize the non-linear prior term via a first-order Taylor expansion around $\mathbf{x}_{0|t}$:
\begin{equation}
    \nabla \Phi(\mathbf{x}) \approx \nabla \Phi(\mathbf{x}_{0|t}) + \mathbf{H}_{\Phi}(\mathbf{x}_{0|t}) (\mathbf{x} - \mathbf{x}_{0|t}),
\end{equation}
where $\mathbf{H}_{\Phi}(\mathbf{x}_{0|t}) \in \mathbb{R}^{N \times N}$ is the Hessian matrix of the prior functional. Substituting this into the optimality condition, we derive the closed-form solution for the optimal image estimate $\bar{\mathbf{x}}_{0|t}$:
\begin{equation}\label{eq:xprior}
    \bar{\mathbf{x}}_{0|t} = \mathbf{x}_{0|t} - \boldsymbol{\Lambda}_t^* \nabla \Phi(\mathbf{x}_{0|t}),
\end{equation}
where the spatially heterogeneous modulation matrix $\boldsymbol{\Lambda}_t^*$ is:
\begin{equation}
    \boldsymbol{\Lambda}_t^* = \left[ \boldsymbol{\Sigma}_t^{-1} + \mathbf{H}_{\Phi}(\mathbf{x}_{0|t}) \right]^{-1} \equiv \boldsymbol{\Sigma}_t \left( \mathbf{I} + \mathbf{H}_{\Phi}(\mathbf{x}_{0|t}) \boldsymbol{\Sigma}_t \right)^{-1}.
\end{equation}
The matrix $\boldsymbol{\Lambda}_t^*$ acts as an adaptive preconditioner, with its modulation strength governed by $\boldsymbol{\Sigma}_t$. When $\boldsymbol{\Sigma}_t \to \mathbf{0}$, we have $\boldsymbol{\Lambda}_t^* \to \mathbf{0}$, which indicates that prior intervention is attenuated to preserve high-frequency details generated by the diffusion model. In contrast, a larger value of $\boldsymbol{\Sigma}_t$ leads to stronger modulation, which enforces more aggressive regularization to suppress structural deviations.

% As $\boldsymbol{\Sigma}_t \to \mathbf{0}$, $\boldsymbol{\Lambda}_t^* \to \mathbf{0}$, meaning the algorithm attenuates the intervention of physical priors to preserve the high-frequency details generated by the diffusion model. Conversely, when $\boldsymbol{\Sigma}_t$ is high, $\boldsymbol{\Lambda}_t^*$ increases accordingly, introducing stronger physical regularization to suppress deviations.

\subsection{State-Aware Trajectory Pruning}
\label{Section4.4}
To alleviate the computational redundancy caused by fixed time steps during the stable evolution phase, LEADer introduces State-Aware Trajectory Pruning (SATP), which incorporates epistemic uncertainty into the temporal dimension to enable adaptive sampling acceleration. In the reverse sampling of diffusion models, a single-step state transition can be approximated via Taylor expansion. Specifically, the state transition equation from time step $t$ to $t - \Delta t$ can be evaluated as:
\begin{equation}
    \mathbf{x}_{t-\Delta t} = \mathbf{x}_t - \Delta t \frac{d\mathbf{x}_t}{dt} + {O}\left( (\Delta t)^2 \left\| \frac{d^2\mathbf{x}_t}{dt^2} \right\| \right).
\end{equation}
The higher-order residual term characterizes the local evolution error, whose magnitude reflects the rate of change in the image state. We relate this quantity to the epistemic uncertainty $\boldsymbol{\Sigma}_t$ derived in Section~\ref{Section4.2}. Specifically, the norm of the second-order derivative is proportional to the trace of $\boldsymbol{\Sigma}_t$:
\begin{equation}
    \left\| \frac{d^2\mathbf{x}_t}{dt^2} \right\| \propto \mathrm{Tr}\left( \mathbf{I} + (1-\bar{\alpha}_t)\nabla_{\mathbf{x}_t}^2 \log p_t(\mathbf{x}_t) \right) \propto \mathrm{Tr}(\boldsymbol{\Sigma}_t) \triangleq {U}_t,
\end{equation}
where $U_t = \frac{1}{N} \text{Tr}\left(\boldsymbol{\Sigma}_t\right)$ denotes the global epistemic uncertainty at time step $t$. To balance reconstruction fidelity and computational efficiency, we introduce a constant information loss budget ${B}$ that constrains the permissible local truncation error. for each step. Specifically, the error induced by a step of size $\Delta t$ is bounded as:
\begin{equation}
    {\delta}(\Delta t) \approx (\Delta t)^\rho \cdot {U}_t \leq {B},
    \label{eq18}
\end{equation}
where $\rho \ge 1$ is the local error order of the sampler. In our implementation, we use the DDIM solver and set $\rho=1$. By solving this error constraint, the maximum admissible step size $\Delta t_{\text{Active}}$ can be derived, and the next evolution state node $t_{\text{Next}}$ can be updated as:
\begin{equation}
\begin{gathered}
\Delta t_{\text{Active}} = \max \left\{1, \lfloor \eta \left( B / (U_t + \boldsymbol{\epsilon}) \right)^{1/\rho} \rfloor \right\}, \\
t_{\text{Next}} = \max \{0, t - \Delta t_{\text{Active}}\},
\end{gathered}
\end{equation}
where $\eta$ is a scaling factor, and $\boldsymbol{\epsilon}$ is a constant that ensures numerical stability. This formulation enables a principled trade-off between efficiency and accuracy. When the model is confident in the current estimate, it indicates that the image is in a stable restoration phase. At this point, the induced local truncation error remains small, thereby permitting larger step sizes and the skipping of redundant iterations. Conversely, in high-uncertainty regions associated with complex structural reconstruction, the step size is reduced to maintain reconstruction accuracy.

\subsection{More Analysis}
\textbf{Algorithm Summary:} In summary, the spatially optimal modulated prior estimate is computed via Eq.\eqref{eq:xprior}.
% \begin{equation}
%     \mathbf{x}_{\text{Prior}} = \mathbf{x}_{0|t} - \boldsymbol{\Lambda}_t^* \nabla \Phi(\boldsymbol{x}_{0|t})
% \end{equation}
To strictly ensure physical measurement consistency, we project it onto the null-space and fuse it with the range-space observation $\mathbf{y}$, which yields the clean image estimate:
\begin{equation}
    \hat{\mathbf{x}}_{0|t} = \mathbf{A}^\dagger \mathbf{y} + (\mathbf{I} - \mathbf{A}^\dagger \mathbf{A}) \bar{\mathbf{x}}_{0|t}.
\end{equation}
Finally, following the non-Markovian formulation of DDIM, the next state $\mathbf{x}_{t_{\text{Next}}}$ can be derived using $\hat{\mathbf{x}}_{0|t}$ and the adaptively selected time step $t_{\text{Next}}$:
\begin{equation}
    \mathbf{x}_{t_{\text{Next}}} = \sqrt{\bar{\alpha}_{t_{\text{Next}}}} \hat{\mathbf{x}}_{0|t} + \sqrt{1 - \bar{\alpha}_{t_{\text{Next}}} - \sigma_{t_{\text{Next}}}^2} \boldsymbol{\epsilon}_\theta(\mathbf{x}_t, t) + \sigma_{t_{\text{Next}}} {\boldsymbol{\epsilon}}_t,
\end{equation}
where ${\boldsymbol{\epsilon}}_t \sim \mathcal{N}(\mathbf{0}, \mathbf{I})$ is the sampling noise. The overall LEADer algorithm is summarized in Algorithm~\ref{Algorithm1}.

\setlength{\algomargin}{1.55em}
\begin{algorithm}[t]
\small
\linespread{1.3}\selectfont
\setlength{\lineskiplimit}{-\maxdimen}
\setlength{\lineskip}{0pt}

\caption{Local Epistemic Uncertainty Guided Active Sampling (LEADer)}
\label{Algorithm1}
\KwIn{Noise estimator $\boldsymbol{\epsilon}_\theta(\cdot, t)$, $T, \mathbf{y}, \mathbf{A}, \Phi, {B}, \eta, \rho, \boldsymbol{\epsilon}$}
\KwOut{Sampled result $\hat{\mathbf{x}}$}

Initialize $\mathbf{x}_T \sim \mathcal{N}(\mathbf{0}, \mathbf{I})$

\While{$t > 0$}{
  \AlgoRow{cyan!10}{\textcolor{cyan!90!black}{\textit{Pre-computation}}}
  
  \AlgoRow{cyan!10}{$\mathbf{x}_{0|t} \leftarrow \frac{1}{\sqrt{\bar{\alpha}_t}} (\mathbf{x}_t - \sqrt{1 - \bar{\alpha}_t} \boldsymbol{\epsilon}_\theta(\mathbf{x}_t, t))$}
  
  \AlgoRow{cyan!10}{$\mathcal{F}_t(\mathbf{x}_t) \leftarrow -\nabla_{\mathbf{x}_t}^2 \log p_t(\mathbf{x}_t)$}
  
  \AlgoRow{cyan!10}{$\boldsymbol{\Sigma}_t \leftarrow \frac{1 - \bar{\alpha}_t}{\bar{\alpha}_t} [\mathbf{I} - (1 - \bar{\alpha}_t) \mathcal{F}_t(\mathbf{x}_t)]$}

  \AlgoRow{teal!10}{\textcolor{teal!90!black}{\textit{Uncertainty-Calibrated Prior Modulation (UCPM)}}}
  
  \AlgoRow{teal!10}{$\boldsymbol{\Lambda}_t^* \leftarrow \boldsymbol{\Sigma}_t (\mathbf{I} + \mathbf{H}_\Phi(\mathbf{x}_{0|t}) \boldsymbol{\Sigma}_t)^{-1}$}
  
  \AlgoRow{teal!10}{$\bar{\mathbf{x}}_{0|t} \leftarrow \mathbf{x}_{0|t} - \boldsymbol{\Lambda}_t^* \nabla \Phi(\mathbf{x}_{0|t})$}
  
  \AlgoRow{teal!10}{$\hat{\mathbf{x}}_{0|t} \leftarrow \mathbf{A}^\dagger \mathbf{y} + (\mathbf{I} - \mathbf{A}^\dagger \mathbf{A}) \bar{\mathbf{x}}_{0|t}$}

  \AlgoRow{orange!10}{\textcolor{orange!90!black}{\textit{State-Aware Trajectory Pruning (SATP)}}}
  
  \AlgoRow{orange!10}{$U_t \leftarrow \frac{1}{N} \text{Tr}(\boldsymbol{\Sigma}_t)$}
  
  \AlgoRow{orange!10}{$\Delta t_{\text{Active}} \leftarrow \max \left\{1, \lfloor \eta \left( B / (U_t + \boldsymbol{\epsilon}) \right)^{1/\rho} \rfloor \right\}$}
  
  \AlgoRow{orange!10}{$t_{\text{Next}} \leftarrow \max \{0, t - \Delta t_{\text{Active}}\}$}

  \AlgoRow{purple!10}{\textcolor{purple!90!black}{\textit{Generative Sampling Step}}}
  
  \AlgoRow{purple!10}{$\boldsymbol{\epsilon}_t \sim \mathcal{N}(\mathbf{0}, \mathbf{I})$}
  
  \AlgoRow{purple!10}{\resizebox{\AlgoRowWidth}{!}{$\mathbf{x}_{t_{\text{Next}}} \leftarrow \sqrt{\bar{\alpha}_{t_{\text{Next}}}} \hat{\mathbf{x}}_{0|t} + \sqrt{1 - \bar{\alpha}_{t_{\text{Next}}} - \sigma_{t_{\text{Next}}}^2} \boldsymbol{\epsilon}_\theta(\mathbf{x}_t, t) + \sigma_{t_{\text{Next}}} \boldsymbol{\epsilon}_t$}}
  
  \AlgoRow{purple!10}{$t \leftarrow t_{\text{Next}}$}
  
}
\KwRet{$\hat{\mathbf{x}}$}
\end{algorithm}

\begin{table*}[!t]
\centering
\caption{Quantitative results for five image restoration tasks on CelebA-HQ 1K (top) and ImageNet 1K (bottom). The Avg. $\Delta$ column reports the average performance improvements and time reductions achieved by our method over the baselines.}
\definecolor{celebours}{HTML}{F5F9D6}
\definecolor{imagenetours}{HTML}{DCF9E3}
\definecolor{psnrpurple}{HTML}{C299FF} 
\definecolor{lpipsblue}{HTML}{66B2FF}  
\definecolor{timegreen}{HTML}{5CD65C}  
\newcommand{\purpledelta}[1]{\textcolor{psnrpurple}{#1}}
\newcommand{\bluedelta}[1]{\textcolor{lpipsblue}{#1}}
\newcommand{\greendelta}[1]{\textcolor{timegreen}{#1}}
\newcommand{\purplebar}[1]{\textcolor{psnrpurple}{\rule{8pt}{#1pt}}}
\newcommand{\bluebar}[1]{\textcolor{lpipsblue}{\rule{8pt}{#1pt}}}
\newcommand{\greenbar}[1]{\textcolor{timegreen}{\rule{8pt}{#1pt}}}
\newcommand{\ccC}{\cellcolor{celebours}}
\newcommand{\ccI}{\cellcolor{imagenetours}}

\resizebox{\textwidth}{!}{
\begin{tabular}{l ccc ccc ccc ccc ccc ccc}
\toprule
\multirow{2.35}{*}{\textbf{Method}} &
\multicolumn{3}{c}{\makecell{Bicub. SR $\times$ 4 \\ ($\sigma_e = 0$)}} &
\multicolumn{3}{c}{\makecell{Bicub. SR $\times$ 4 \\ ($\sigma_e = 0.05$)}} &
\multicolumn{3}{c}{\makecell{Gaussian Deb.  \\ ($\sigma_e = 0.05$)}} &
\multicolumn{3}{c}{\makecell{Motion Deb.  \\ ($\sigma_e = 0.05$)}} &
\multicolumn{3}{c}{\makecell{CS 25\% \\ ($\sigma_e = 0.05$)}} &
\multicolumn{3}{c}{\textbf{Avg. $\Delta$}} \\
\cmidrule(lr){2-4} \cmidrule(lr){5-7} \cmidrule(lr){8-10} \cmidrule(lr){11-13} \cmidrule(lr){14-16} \cmidrule(lr){17-19}
 & PSNR $\uparrow$ & LPIPS $\downarrow$ & Time $\downarrow$ & PSNR $\uparrow$ & LPIPS $\downarrow$ & Time $\downarrow$ & PSNR $\uparrow$ & LPIPS $\downarrow$ & Time $\downarrow$ & PSNR $\uparrow$ & LPIPS $\downarrow$ & Time $\downarrow$ & PSNR $\uparrow$ & LPIPS $\downarrow$ & Time $\downarrow$ & PSNR $\uparrow$ & LPIPS $\downarrow$ & Time $\downarrow$ \\
\midrule
\multicolumn{19}{c}{\textbf{CelebA-HQ 1K}} \\
\midrule
DDRM~\cite{DBLP:conf/nips/KawarEES22} \textcolor{gray}{\scriptsize [NeurLPS2023]}
& $31.64_{\pm2.43}$ & $0.054_{\pm0.021}$ & 3.87 s
& $29.26_{\pm1.86}$ & $0.090_{\pm0.028}$ & 4.05 s
& $30.53_{\pm1.74}$ & $0.074_{\pm0.024}$ & 4.06 s
& N / A & N / A & N / A
& $26.07_{\pm2.08}$ & $0.128_{\pm0.037}$ & 3.86 s
& N / A & N / A & N / A \\
DPS~\cite{DBLP:conf/iclr/ChungKMKY23} \textcolor{gray}{\scriptsize [ICLR2023]}
& $29.39_{\pm3.16}$ & $0.065_{\pm0.026}$ & 134.20 s
& $27.49_{\pm2.84}$ & $0.086_{\pm0.032}$ & 138.25 s
& $27.75_{\pm2.57}$ & $0.084_{\pm0.029}$ & 143.69 s
& $19.63_{\pm3.41}$ & $0.227_{\pm0.043}$ & 148.69 s
& N / A & N / A & N / A
& N / A & N / A & N / A \\
DiffPIR~\cite{DBLP:conf/cvpr/ZhuZLCWTG23} \textcolor{gray}{\scriptsize [CVPR2023]}
& $30.36_{\pm2.27}$ & $0.051_{\pm0.019}$ & 7.45 s
& $27.44_{\pm1.93}$ & $0.085_{\pm0.027}$ & 7.68 s
& $28.89_{\pm1.78}$ & $0.074_{\pm0.022}$ & 7.66 s
& $27.96_{\pm1.61}$ & $0.102_{\pm0.031}$ & 7.58 s
& N / A & N / A & N / A
& N / A & N / A & N / A  \\
IDPG~\cite{DBLP:conf/cvpr/GarberT24} \textcolor{gray}{\scriptsize [CVPR2024]}
& $32.66_{\pm2.18}$ & $0.111_{\pm0.023}$ & 4.75 s
& $29.89_{\pm1.64}$ & $0.155_{\pm0.031}$ & 4.88 s
& $31.08_{\pm1.52}$ & $0.150_{\pm0.027}$ & 4.81 s
& $29.73_{\pm1.73}$ & $0.134_{\pm0.025}$ & 5.59 s
& $26.26_{\pm1.95}$ & $0.143_{\pm0.033}$ & 5.00 s
& N / A & N / A & N / A \\
% PIRP F$_{\text{lb}}\text{G}_{+}$~\cite{DBLP:journals/kbs/YangZZZ26} \textcolor{gray}{\scriptsize [KBS2026]}
% & $33.06_{\pm1.94}$ & $0.116_{\pm0.021}$ & 4.78 s
% & $30.14_{\pm1.48}$ & $0.161_{\pm0.029}$ & 4.92 s
% & $31.68_{\pm1.39}$ & $0.140_{\pm0.024}$ & 4.85 s
% & $30.28_{\pm1.66}$ & $0.144_{\pm0.026}$ & 5.61 s
% & $26.62_{\pm1.82}$ & $0.143_{\pm0.031}$ & 5.03 s
% & N / A & N / A & N / A \\

\hdashline
DDNM~\cite{DBLP:conf/iclr/WangYZ23} \textcolor{gray}{\scriptsize [ICLR2023]} & 31.64$_{\pm3.87}$ & 0.048$_{\pm0.024}$ & 3.77 s & 29.19$_{\pm2.51}$ & 0.082$_{\pm0.031}$ & 3.87 s & 30.60$_{\pm2.02}$ & 0.072$_{\pm0.022}$ & 3.90 s & N / A & N / A & N / A & 26.14$_{\pm2.18}$ & 0.124$_{\pm0.043}$ & 3.83 s &
\multirow{3}{*}{\begin{tabular}[t]{@{}c@{}}\purpledelta{\textbf{$+$2.382\%}}\\[2pt]\purplebar{11.9}\end{tabular}} &
\multirow{3}{*}{\begin{tabular}[t]{@{}c@{}}\bluedelta{\textbf{$-$12.781\%}}\\[2pt]\bluebar{20.2}\end{tabular}} &
\multirow{3}{*}{\begin{tabular}[t]{@{}c@{}}\greendelta{\textbf{$-$5.747\%}}\\[2pt]\greenbar{9.6}\end{tabular}} \\
+ EquS$^+$~\cite{DBLP:journals/corr/abs-2511-09965} \textcolor{gray}{\scriptsize [WACV2026]} & 32.44$_{\pm2.58}$ & 0.053$_{\pm0.019}$ & 3.83 s & 29.20$_{\pm1.38}$ & 0.093$_{\pm0.027}$ & 3.87 s & 30.83$_{\pm1.56}$ & 0.069$_{\pm0.021}$ & 3.96 s & N / A & N / A & N / A & 26.31$_{\pm2.45}$ & 0.119$_{\pm0.035}$ & 4.02 s & & & \\
\ccC + LEADer \textcolor{gray}{\scriptsize [Ours]} & \ccC 32.73$_{\pm2.12}$ & \ccC 0.044$_{\pm0.017}$ & \ccC 3.42 s & \ccC 29.69$_{\pm1.10}$ & \ccC 0.077$_{\pm0.021}$ & \ccC 3.63 s & \ccC 30.86$_{\pm1.38}$ & \ccC 0.063$_{\pm0.015}$ & \ccC 3.75  s& \ccC N / A & \ccC N / A & \ccC N / A & \ccC 27.06$_{\pm1.42}$ & \ccC 0.094$_{\pm0.023}$ & \ccC 3.69 s & & & \\

\hdashline
DDPG~\cite{DBLP:conf/cvpr/GarberT24} \textcolor{gray}{\scriptsize [CVPR2024]}
& $31.60_{\pm2.36}$ & $0.052_{\pm0.021}$ & 4.96 s
& $29.39_{\pm1.74}$ & $0.105_{\pm0.029}$ & 5.09 s
& $30.41_{\pm1.58}$ & $0.068_{\pm0.023}$ & 5.11 s
& $29.02_{\pm1.69}$ & $0.082_{\pm0.026}$ & 5.72 s
& $26.76_{\pm1.95}$ & $0.152_{\pm0.034}$ & 5.24 s &
\multirow{3}{*}{\begin{tabular}[t]{@{}c@{}}\purpledelta{\textbf{$+$1.197\%}}\\[2pt]\purplebar{6.0}\end{tabular}} &
\multirow{3}{*}{\begin{tabular}[t]{@{}c@{}}\bluedelta{\textbf{$-$7.137\%}}\\[2pt]\bluebar{11.7}\end{tabular}} &
\multirow{3}{*}{\begin{tabular}[t]{@{}c@{}}\greendelta{\textbf{$-$6.662\%}}\\[2pt]\greenbar{11.0}\end{tabular}} \\
+ EquS~\cite{DBLP:journals/corr/abs-2511-09965} \textcolor{gray}{\scriptsize [WACV2026]}
& $31.73_{\pm2.08}$ & $0.054_{\pm0.019}$ & 4.88 s
& $29.52_{\pm1.53}$ & $0.109_{\pm0.027}$ & 5.01 s
& $30.47_{\pm1.42}$ & $0.071_{\pm0.021}$ & 5.03 s
& $29.17_{\pm1.55}$ & $0.088_{\pm0.024}$ & 5.62 s
& $26.82_{\pm1.76}$ & $0.155_{\pm0.032}$ & 5.15 s & & & \\
\ccC + LEADer \textcolor{gray}{\scriptsize [Ours]} & \ccC $31.94_{\pm1.79}$ & \ccC $0.047_{\pm0.017}$ & \ccC 4.63 s & \ccC $29.65_{\pm1.31}$ & \ccC $0.098_{\pm0.023}$ & \ccC 4.75 s & \ccC $30.77_{\pm1.24}$ & \ccC $0.062_{\pm0.018}$ & \ccC 4.77 s & \ccC $29.40_{\pm1.37}$ & \ccC $0.076_{\pm0.021}$ & \ccC 5.34 s & \ccC $27.17_{\pm1.58}$ & \ccC $0.144_{\pm0.028}$ & \ccC 4.89 s & & & \\

\hdashline
ProjDiff~\cite{DBLP:conf/nips/ZhangZJLG24} \textcolor{gray}{\scriptsize [NeurIPS2024]}
& $32.57_{\pm1.88}$ & $0.091_{\pm0.022}$ & 3.94 s
& $29.49_{\pm1.42}$ & $0.080_{\pm0.024}$ & 4.12 s
& $31.41_{\pm1.35}$ & $0.068_{\pm0.020}$ & 4.27 s
& N / A & N / A & N / A
& $26.41_{\pm1.57}$ & $0.132_{\pm0.031}$ & 4.05 s &
\multirow{3}{*}{\begin{tabular}[t]{@{}c@{}}\purpledelta{\textbf{$+$0.882\%}}\\[2pt]\purplebar{4.4}\end{tabular}} &
\multirow{3}{*}{\begin{tabular}[t]{@{}c@{}}\bluedelta{\textbf{$-$7.330\%}}\\[2pt]\bluebar{12.0}\end{tabular}} &
\multirow{3}{*}{\begin{tabular}[t]{@{}c@{}}\greendelta{\textbf{$-$8.421\%}}\\[2pt]\greenbar{13.6}\end{tabular}} \\
+ EquS$^+$~\cite{DBLP:journals/corr/abs-2511-09965} \textcolor{gray}{\scriptsize [WACV2026]}
& $32.64_{\pm1.63}$ & $0.094_{\pm0.020}$ & 3.87 s
& $29.53_{\pm1.21}$ & $0.080_{\pm0.021}$ & 4.04 s
& $31.45_{\pm1.18}$ & $0.071_{\pm0.018}$ & 4.18 s
& N / A & N / A & N / A
& $26.45_{\pm1.39}$ & $0.135_{\pm0.028}$ & 3.97 s & & & \\
\ccC + LEADer \textcolor{gray}{\scriptsize [Ours]} & \ccC $32.86_{\pm1.34}$ & \ccC $0.084_{\pm0.017}$ & \ccC 3.61 s & \ccC $29.75_{\pm1.05}$ & \ccC $0.074_{\pm0.018}$ & \ccC 3.77 s & \ccC $31.64_{\pm0.98}$ & \ccC $0.062_{\pm0.015}$ & \ccC 3.90 s & \ccC N / A & \ccC N / A & \ccC N / A & \ccC $26.68_{\pm1.16}$ & \ccC $0.125_{\pm0.023}$ & \ccC 3.72 s & & & \\

\hdashline
PIRP F$_{\text{lb}}\text{G}_{-}$~\cite{DBLP:journals/kbs/YangZZZ26} \textcolor{gray}{\scriptsize [KBS2026]}
& $30.80_{\pm2.11}$ & $0.042_{\pm0.018}$ & 5.01 s
& $28.55_{\pm1.57}$ & $0.079_{\pm0.024}$ & 5.14 s
& $29.90_{\pm1.46}$ & $0.061_{\pm0.020}$ & 5.12 s
& $28.50_{\pm1.63}$ & $0.074_{\pm0.022}$ & 5.77 s
& $26.59_{\pm1.74}$ & $0.091_{\pm0.027}$ & 5.22 s &
\multirow{3}{*}{\begin{tabular}[t]{@{}c@{}}\purpledelta{\textbf{$+$0.934\%}}\\[2pt]\purplebar{4.7}\end{tabular}} &
\multirow{3}{*}{\begin{tabular}[t]{@{}c@{}}\bluedelta{\textbf{$-$7.618\%}}\\[2pt]\bluebar{12.4}\end{tabular}} &
\multirow{3}{*}{\begin{tabular}[t]{@{}c@{}}\greendelta{\textbf{$-$5.824\%}}\\[2pt]\greenbar{9.7}\end{tabular}} \\
+ EquS~\cite{DBLP:journals/corr/abs-2511-09965} \textcolor{gray}{\scriptsize [WACV2026]}
& $30.85_{\pm1.86}$ & $0.043_{\pm0.016}$ & 4.94 s
& $28.60_{\pm1.39}$ & $0.078_{\pm0.021}$ & 5.06 s
& $29.94_{\pm1.28}$ & $0.062_{\pm0.018}$ & 5.04 s
& $28.55_{\pm1.47}$ & $0.075_{\pm0.020}$ & 5.67 s
& $26.63_{\pm1.58}$ & $0.092_{\pm0.024}$ & 5.14 s & & & \\
\ccC + LEADer \textcolor{gray}{\scriptsize [Ours]} & \ccC $31.02_{\pm1.61}$ & \ccC $0.039_{\pm0.014}$ & \ccC 4.72 s & \ccC $28.86_{\pm1.18}$ & \ccC $0.072_{\pm0.018}$ & \ccC 4.84 s & \ccC $30.16_{\pm1.09}$ & \ccC $0.055_{\pm0.015}$ & \ccC 4.83 s & \ccC $28.77_{\pm1.26}$ & \ccC $0.069_{\pm0.017}$ & \ccC 5.43 s & \ccC $26.87_{\pm1.37}$ & \ccC $0.086_{\pm0.021}$ & \ccC 4.91 s & & & \\

\bottomrule
\multicolumn{19}{c}{\textbf{ImageNet 1K}} \\
\midrule
DDRM~\cite{DBLP:conf/nips/KawarEES22} \textcolor{gray}{\scriptsize [NeurLPS2023]} & 27.38$_{\pm2.12}$ & 0.270$_{\pm0.045}$ & 8.16 s & 25.54$_{\pm1.52}$ & 0.333$_{\pm0.047}$ & 8.27 s & 27.71$_{\pm1.83}$ & 0.243$_{\pm0.032}$ & 8.02 s & N / A & N / A & N / A & 21.58$_{\pm1.75}$ & 0.301$_{\pm0.038}$ & 8.07 s & N / A & N / A & N / A \\
DPS~\cite{DBLP:conf/iclr/ChungKMKY23} \textcolor{gray}{\scriptsize [ICLR2023]} & 25.56$_{\pm1.85}$ & 0.236$_{\pm0.029}$ & 246.14 s & 24.05$_{\pm1.43}$ & 0.271$_{\pm0.045}$ & 263.95 s & 26.64$_{\pm1.57}$ & 0.240$_{\pm0.032}$ & 248.26 s & 17.52$_{\pm1.78}$ & 0.468$_{\pm0.066}$ & 275.41 s & N / A & N / A & N / A & N / A & N / A & N / A \\
DiffPIR~\cite{DBLP:conf/cvpr/ZhuZLCWTG23} \textcolor{gray}{\scriptsize [CVPR2023]} & 26.99$_{\pm1.92}$ & 0.255$_{\pm0.033}$ & 15.40 s & 24.65$_{\pm1.32}$ & 0.318$_{\pm0.044}$ & 15.62 s & 26.64$_{\pm1.74}$ & 0.240$_{\pm0.029}$ & 15.49 s & 25.34$_{\pm1.52}$ & 0.284$_{\pm0.038}$ & 18.19 s & N / A & N / A & N / A & N / A & N / A & N / A \\
IDPG~\cite{DBLP:conf/cvpr/GarberT24} \textcolor{gray}{\scriptsize [CVPR2024]} & 27.20$_{\pm1.68}$ & 0.326$_{\pm0.047}$ & 8.42 s & 25.51$_{\pm1.22}$ & 0.411$_{\pm0.054}$ & 8.53 s & 27.47$_{\pm1.83}$ & 0.313$_{\pm0.045}$ & 8.46 s & 26.02$_{\pm1.56}$ & 0.354$_{\pm0.048}$ & 9.77 s & 22.37$_{\pm1.29}$ & 0.261$_{\pm0.033}$ & 8.48 s & N / A & N / A & N / A \\
ProjDiff~\cite{DBLP:conf/nips/ZhangZJLG24} \textcolor{gray}{\scriptsize [NeurIPS2024]} & 27.09$_{\pm1.53}$ & 0.242$_{\pm0.031}$ & 8.13 s & 25.73$_{\pm1.11}$ & 0.336$_{\pm0.046}$ & 8.15 s & 27.91$_{\pm1.78}$ & 0.238$_{\pm0.029}$ & 8.11 s & N / A & N / A & N / A & 22.06$_{\pm1.26}$ & 0.248$_{\pm0.033}$ & 8.32 s & N / A & N / A & N / A \\
% PIRP F$_{\text{lb}}\text{G}_{+}$~\cite{DBLP:journals/kbs/YangZZZ26} \textcolor{gray}{\scriptsize [KBS2026]} & 27.41$_{\pm1.85}$ & 0.255$_{\pm0.034}$ & 8.46 s & 25.56$_{\pm1.39}$ & 0.356$_{\pm0.052}$ & 8.58 s & 27.75$_{\pm1.72}$ & 0.209$_{\pm0.026}$ & 8.51 s & 26.00$_{\pm1.48}$ & 0.360$_{\pm0.050}$ & 9.86 s & 22.54$_{\pm1.31}$ & 0.257$_{\pm0.031}$ & 8.17 s & N / A & N / A & N / A \\

\hdashline
DDNM~\cite{DBLP:conf/iclr/WangYZ23} \textcolor{gray}{\scriptsize [ICLR2023]} & 27.45$_{\pm2.12}$ & 0.245$_{\pm0.035}$ & 8.03 s & 25.52$_{\pm1.48}$ & 0.324$_{\pm0.047}$ & 8.06 s & 27.69$_{\pm2.05}$ & 0.237$_{\pm0.027}$ & 7.97 s & N / A & N / A & N / A & 21.66$_{\pm1.35}$ & 0.265$_{\pm0.041}$ & 8.20 s &
\multirow{3}{*}{\begin{tabular}[t]{@{}c@{}}\purpledelta{\textbf{$+$2.623\%}}\\[2pt]\purplebar{13.1}\end{tabular}} &
\multirow{3}{*}{\begin{tabular}[t]{@{}c@{}}\bluedelta{\textbf{$-$9.944\%}}\\[2pt]\bluebar{15.9}\end{tabular}} &
\multirow{3}{*}{\begin{tabular}[t]{@{}c@{}}\greendelta{\textbf{$-$3.044\%}}\\[2pt]\greenbar{5.6}\end{tabular}} \\
+ EquS$^+$~\cite{DBLP:journals/corr/abs-2511-09965} \textcolor{gray}{\scriptsize [WACV2026]} & 27.63$_{\pm2.03}$ & 0.242$_{\pm0.033}$ & 8.00 s & 25.56$_{\pm1.37}$ & 0.327$_{\pm0.045}$ & 8.12 s & 27.72$_{\pm1.87}$ & 0.234$_{\pm0.025}$ & 8.01 s & N / A & N / A & N / A & 22.14$_{\pm1.25}$ & 0.246$_{\pm0.038}$ & 8.21 s & & & \\
\ccI + LEADer \textcolor{gray}{\scriptsize [Ours]} & \ccI 27.71$_{\pm1.92}$ & \ccI 0.231$_{\pm0.029}$ & \ccI 7.74 s & \ccI 25.63$_{\pm1.28}$ & \ccI 0.306$_{\pm0.038}$ & \ccI 7.86 s & \ccI 27.81$_{\pm1.65}$ & \ccI 0.207$_{\pm0.022}$ & \ccI 7.66 s & \ccI N / A & \ccI N / A & \ccI N / A & \ccI 23.54$_{\pm1.36}$ & \ccI 0.223$_{\pm0.030}$ & \ccI 8.02 s & & & \\

\hdashline
DDPG~\cite{DBLP:conf/cvpr/GarberT24} \textcolor{gray}{\scriptsize [CVPR2024]}
& $27.41_{\pm1.86}$ & $0.255_{\pm0.031}$ & 8.43 s
& $25.55_{\pm1.42}$ & $0.354_{\pm0.044}$ & 8.17 s
& $27.73_{\pm1.74}$ & $0.205_{\pm0.026}$ & 8.24 s
& $25.94_{\pm1.58}$ & $0.249_{\pm0.033}$ & 8.79 s
& $20.69_{\pm1.27}$ & $0.258_{\pm0.037}$ & 8.56 s &
\multirow{3}{*}{\begin{tabular}[t]{@{}c@{}}\purpledelta{\textbf{$+$1.079\%}}\\[2pt]\purplebar{5.4}\end{tabular}} &
\multirow{3}{*}{\begin{tabular}[t]{@{}c@{}}\bluedelta{\textbf{$-$4.914\%}}\\[2pt]\bluebar{8.4}\end{tabular}} &
\multirow{3}{*}{\begin{tabular}[t]{@{}c@{}}\greendelta{\textbf{$-$3.169\%}}\\[2pt]\greenbar{5.8}\end{tabular}} \\
+ EquS~\cite{DBLP:journals/corr/abs-2511-09965} \textcolor{gray}{\scriptsize [WACV2026]}
& $27.44_{\pm1.73}$ & $0.252_{\pm0.029}$ & 8.49 s
& $25.63_{\pm1.35}$ & $0.351_{\pm0.041}$ & 8.23 s
& $27.73_{\pm1.61}$ & $0.203_{\pm0.024}$ & 8.31 s
& $25.99_{\pm1.46}$ & $0.247_{\pm0.031}$ & 8.86 s
& $20.71_{\pm1.18}$ & $0.255_{\pm0.034}$ & 8.63 s & & & \\
\ccI + LEADer \textcolor{gray}{\scriptsize [Ours]} & \ccI $27.62_{\pm1.52}$ & \ccI $0.243_{\pm0.024}$ & \ccI 8.18 s & \ccI $25.75_{\pm1.21}$ & \ccI $0.339_{\pm0.036}$ & \ccI 7.95 s & \ccI $28.02_{\pm1.43}$ & \ccI $0.194_{\pm0.021}$ & \ccI 8.00 s & \ccI $26.19_{\pm1.33}$ & \ccI $0.236_{\pm0.028}$ & \ccI 8.51 s & \ccI $21.07_{\pm1.09}$ & \ccI $0.245_{\pm0.030}$ & \ccI 8.21 s & & & \\

\hdashline
SITCOM~\cite{DBLP:conf/icml/AlkhouriLHD0RW25} \textcolor{gray}{\scriptsize [ICML2025]} & 27.26$_{\pm1.32}$ & 0.186$_{\pm0.026}$ & 154.64 s & 25.35$_{\pm1.21}$ & 0.232$_{\pm0.038}$ & 163.50 s & $27.40_{\pm0.45}$ & $0.236_{\pm0.039}$ & 142.36 s & $28.65_{\pm0.34}$ & $0.189_{\pm0.036}$ & 140.87 s & N / A & N / A & N / A &
\multirow{3}{*}{\begin{tabular}[t]{@{}c@{}}\purpledelta{\textbf{$+$0.809\%}}\\[2pt]\purplebar{4.0}\end{tabular}} &
\multirow{3}{*}{\begin{tabular}[t]{@{}c@{}}\bluedelta{\textbf{$-$4.828\%}}\\[2pt]\bluebar{8.2}\end{tabular}} &
\multirow{3}{*}{\begin{tabular}[t]{@{}c@{}}\greendelta{\textbf{$-$5.470\%}}\\[2pt]\greenbar{9.2}\end{tabular}} \\
+ EquS~\cite{DBLP:journals/corr/abs-2511-09965} \textcolor{gray}{\scriptsize [WACV2026]} & 27.30$_{\pm1.30}$ & 0.181$_{\pm0.024}$ & 151.88 s & 25.36$_{\pm1.19}$ & 0.228$_{\pm0.036}$ & 160.42 s & $27.38_{\pm0.44}$ & $0.232_{\pm0.037}$ & 139.93 s & $28.66_{\pm0.33}$ & $0.185_{\pm0.034}$ & 138.51 s & N / A & N / A & N / A & & & \\
\ccI + LEADer \textcolor{gray}{\scriptsize [Ours]} & \ccI 27.56$_{\pm1.28}$ & \ccI 0.176$_{\pm0.022}$ & \ccI 145.97 s & \ccI 25.53$_{\pm1.16}$ & \ccI 0.223$_{\pm0.033}$ & \ccI 154.36 s & \ccI $27.59_{\pm0.43}$ & \ccI $0.226_{\pm0.035}$ & \ccI 134.68 s & \ccI $28.86_{\pm0.32}$ & \ccI $0.178_{\pm0.031}$ & \ccI 133.42 s & \ccI N / A & \ccI N / A & \ccI N / A & & & \\

\hdashline
PIRP F$_{\text{lb}}\text{G}_{-}$~\cite{DBLP:journals/kbs/YangZZZ26} \textcolor{gray}{\scriptsize [KBS2026]}
& $24.86_{\pm1.34}$ & $0.151_{\pm0.024}$ & 8.55 s
& $25.07_{\pm1.52}$ & $0.290_{\pm0.041}$ & 8.56 s
& $26.17_{\pm1.28}$ & $0.150_{\pm0.022}$ & 8.69 s
& $24.81_{\pm1.67}$ & $0.209_{\pm0.035}$ & 9.83 s
& $23.68_{\pm1.45}$ & $0.217_{\pm0.031}$ & 8.16 s &
\multirow{3}{*}{\begin{tabular}[t]{@{}c@{}}\purpledelta{\textbf{$+$1.487\%}}\\[2pt]\purplebar{7.4}\end{tabular}} &
\multirow{3}{*}{\begin{tabular}[t]{@{}c@{}}\bluedelta{\textbf{$-$2.726\%}}\\[2pt]\bluebar{5.1}\end{tabular}} &
\multirow{3}{*}{\begin{tabular}[t]{@{}c@{}}\greendelta{\textbf{$-$4.010\%}}\\[2pt]\greenbar{7.0}\end{tabular}} \\
+ EquS~\cite{DBLP:journals/corr/abs-2511-09965} \textcolor{gray}{\scriptsize [WACV2026]}
& $24.91_{\pm1.21}$ & $0.153_{\pm0.021}$ & 8.58 s
& $25.02_{\pm1.39}$ & $0.294_{\pm0.038}$ & 8.61 s
& $26.11_{\pm1.16}$ & $0.155_{\pm0.020}$ & 8.73 s
& $24.76_{\pm1.48}$ & $0.211_{\pm0.032}$ & 9.88 s
& $23.71_{\pm1.33}$ & $0.220_{\pm0.029}$ & 8.19 s & & & \\
\ccI + LEADer \textcolor{gray}{\scriptsize [Ours]} & \ccI $25.37_{\pm1.05}$ & \ccI $0.146_{\pm0.018}$ & \ccI 8.29 s & \ccI $25.34_{\pm1.12}$ & \ccI $0.278_{\pm0.030}$ & \ccI 8.32 s & \ccI $26.28_{\pm0.98}$ & \ccI $0.152_{\pm0.019}$ & \ccI 8.41 s & \ccI $25.24_{\pm1.21}$ & \ccI $0.201_{\pm0.028}$ & \ccI 9.63 s & \ccI $24.19_{\pm1.10}$ & \ccI $0.209_{\pm0.026}$ & \ccI 7.43 s & & & \\

\bottomrule
\end{tabular}
\label{Table1}
}
\vspace{-3mm}
\end{table*}

\noindent\textbf{Plug-and-Play Analysis:} 
Notably, the LEADer framework can be seamlessly integrated into existing baseline sampling methods as a plug-and-play module. The computations of UCPM and SATP rely entirely on the posterior statistics ($\mathbf{x}_{0|t}$ and $\boldsymbol{\Sigma}_t$) of the diffusion model during the pure inference phase. This eliminates the need for any degradation-specific fine-tuning or retraining of the pre-trained denoising network $\boldsymbol{\epsilon}_\theta$. This design ensures broad compatibility with existing zero-shot image restoration frameworks, facilitating its application across diverse inverse imaging tasks without modifying the underlying model parameters. %, thereby simultaneously enhancing the restoration performance and sampling efficiency across various baseline methods.

\noindent\textbf{Theoretical Analysis:} We now provide theoretical insights into the behavior of the UCPM and SATP mechanisms within the LEADer framework. In particular, we analyze how UCPM maintains data fidelity while introducing spatial prior modulation (Proposition 4.1), and how SATP guarantees a bounded global evolution error during adaptive temporal acceleration (Proposition 4.2). The proofs of the two Propositions are provided in the Supplementary Material.

\noindent \textsc{Proposition 4.1.} \textit{For a linear degradation operator $\mathbf{A}$ and an observation $\mathbf{y}$, let the null-space projection be ${P}_N = \mathbf{I} - \mathbf{A}^\dagger \mathbf{A}$. Since the modulated estimation of UCPM is defined as $\hat{\mathbf{x}} = \mathbf{A}^\dagger \mathbf{y} + {P}_N \mathbf{x}_{Prior}$, under the noise-free assumption, the final generated image strictly satisfies the data fidelity condition:}
\begin{equation}
    \| \mathbf{y} - \mathbf{A}\hat{\mathbf{x}} \|_2^2 = 0.
\end{equation}
Since $\mathbf{A}{P}_N \equiv \mathbf{0}$, all prior modulation induced by the epistemic uncertainty $\boldsymbol{\Sigma}_t$ is strictly constrained within the null space. This implies that UCPM does not interfere with the known range-space components when suppressing artifacts or synthesizing details, thereby circumventing the common issue of data fidelity corruption caused by prior-driven updates.

\noindent \textsc{Proposition 4.2.} \textit{Let $\mathbf{x}^{Dense}$ denote the continuous exact solution of the ODE and $\mathbf{x}^{LEADer}$ the discrete solution accelerated by SATP. Assuming the drift field is $L$-Lipschitz continuous and SATP bounds the single-step truncation error within the budget ${B}$. Based on Gr\"onwall's inequality, the global accumulated error possesses a deterministic analytical upper bound:}
\begin{equation}
    \| \mathbf{x}^{Dense} - \mathbf{x}^{LEADer} \|_2 \le C \cdot {B} \cdot \frac{e^{LT}-1}{L},
\end{equation}
\textit{where $C > 0$ is a constant, $T$ is the total samplings steps of the reverse process, and $L$ is the Lipschitz constant.}

In the discrete ODE solution process, due to Lipschitz continuity, the local truncation error accumulates exponentially over time steps, yielding the $e^{LT}$ term. By introducing the uncertainty trace $U_t$ to adjust the sampling step size, SATP constrains the local error at each step within the predefined tolerance ${B}$. This state-aware local error control ensures that the global accumulated error remains bounded under non-uniform sampling, thereby guaranteeing the convergence of the accelerated generation trajectory.

% The proofs of Propositions 4.1 and 4.2 are in the Appendix.

\begin{figure*}[!t]
\centering
\includegraphics[width=0.98\textwidth]{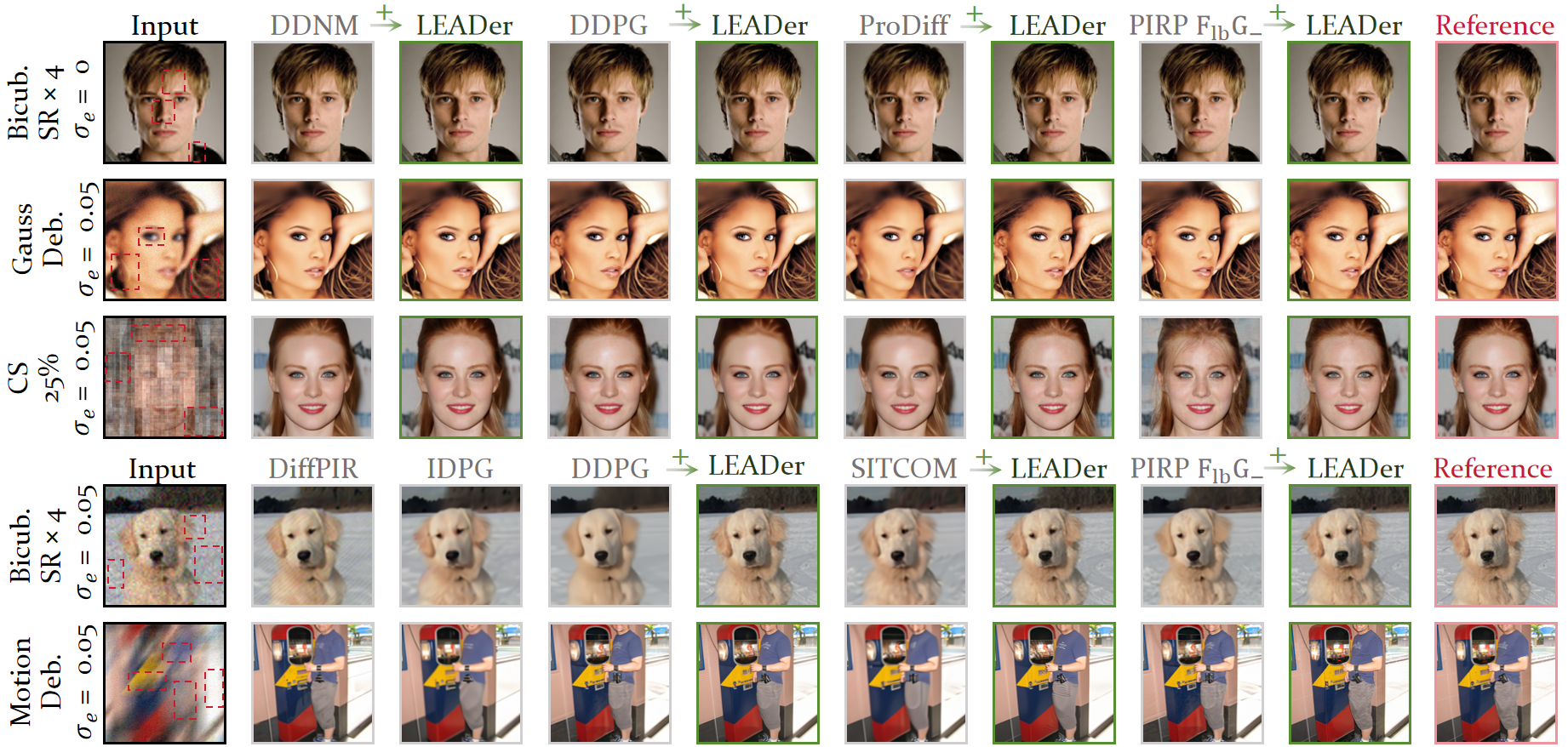} 
\caption{Qualitative comparisons on five image restoration tasks.}
\label{Figure3}
\end{figure*}

\begin{figure*}[!t]
\centering
\includegraphics[width=0.98\textwidth]{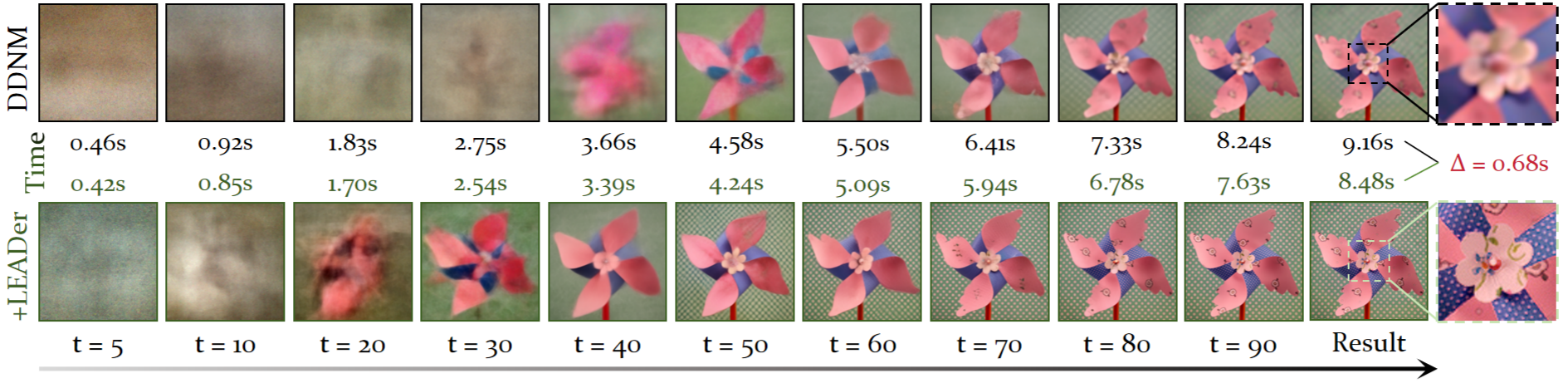} 
\vspace{-4mm}
\caption{Comparison of $\hat{\mathbf{x}}_{0|t}$ between DDNM and LEADer at different NFEs.}
\label{Figure4}
\vspace{-2.5mm}
\end{figure*}

\section{Experiments}
\subsection{Experimental Setup}
We conduct experiments on five typical IR tasks: $4\times$ Super-Resolution (SR) with a bicubic downsampler ($\sigma_e = 0$ and $\sigma_e = 0.05$), Gaussian deblurring ($\sigma_e = 0.05$), motion deblurring ($\sigma_e = 0.05$), and Compressed Sensing (CS) using a Walsh-Hadamard sampling matrix with a 0.25 compression ratio. For validation, we use two standard image restoration benchmarks, i.e., CelebA-HQ 1K~\citep{DBLP:conf/iccv/LiuLWT15} and ImageNet 1K~\citep{DBLP:journals/ijcv/RussakovskyDSKS15}, both at $256\times256$ resolution. For a fair comparison, all methods utilize the same pre-trained DDM denoisers, i.e., one trained on CelebA-HQ~\citep{DBLP:conf/cvpr/LugmayrDRYTG22} and one trained on ImageNet~\citep{DBLP:conf/nips/DhariwalN21}. Sampling steps are uniformly set to $T = 100$ for all methods, except for DPS~\cite{DBLP:conf/iclr/ChungKMKY23} ($T = 1000$) and SITCOM~\cite{DBLP:conf/icml/AlkhouriLHD0RW25}. Note that due to the step-skipping capability of SATP, our total Number of Function Evaluations (NFEs) is less than 100. All experiments were performed on an NVIDIA GeForce RTX 3090 GPU.

\noindent\textbf{Evaluation Metrics.} We employ three prevalent metrics to evaluate the proposed method on image restoration tasks, which include Peak Signal-to-Noise Ratio (PSNR) for distortion evaluation, Learned Perceptual Image Patch Similarity (LPIPS)~\citep{DBLP:conf/cvpr/ZhangIESW18} for perceptual quality, and average per-image Run-time $\left(\text{Time}\right)$ for computational efficiency. We additionally report memory consumption to assess resource usage. Standard deviations of PSNR and LPIPS are provided to reflect performance stability.

\noindent\textbf{Comparison with Other Methods.} To evaluate the effectiveness of LEADer, we integrate it as a plug-and-play module into several advanced zero-shot DMIR approaches, including DDNM~\cite{DBLP:conf/iclr/WangYZ23}, IDPG~\cite{DBLP:conf/cvpr/GarberT24}, DDPG~\cite{DBLP:conf/cvpr/GarberT24}, ProjDiff~\cite{DBLP:conf/nips/ZhangZJLG24}, SITCOM~\cite{DBLP:conf/icml/AlkhouriLHD0RW25}, and PIRP~\cite{DBLP:journals/kbs/YangZZZ26}. We apply the same integration for EquS~\cite{DBLP:journals/corr/abs-2511-09965} and compare their respective effects on quality and efficiency across these baselines. Additionally, we compare LEADer against other SOTA zero-shot DMIR methods, including DDRM~\cite{DBLP:conf/nips/KawarEES22}, DPS~\cite{DBLP:conf/iclr/ChungKMKY23} and DiffPIR~\cite{DBLP:conf/cvpr/ZhuZLCWTG23}.

\subsection{Improvements to DMIR Methods}
\noindent\textbf{Quantitative Results.} To validate the effectiveness of LEADer as a plug-and-play method for DMIR, we conduct detailed quantitative evaluations across five image restoration tasks on CelebA-HQ 1K and ImageNet 1K. As shown in Table~\ref{Table1}, experimental results confirm the superior performance of LEADer. When integrated into various zero-shot DMIR methods, such as DDNM~\cite{DBLP:conf/iclr/WangYZ23}, DDPG~\cite{DBLP:conf/cvpr/GarberT24}, ProjDiff~\cite{DBLP:conf/nips/ZhangZJLG24}, SITCOM~\cite{DBLP:conf/icml/AlkhouriLHD0RW25}, and PIRP~\cite{DBLP:journals/kbs/YangZZZ26}, LEADer consistently enhances restoration quality. Specifically, on CelebA-HQ 1K, LEADer achieves average PSNR gains ranging from 0.88\% to 2.38\%, while significantly reducing LPIPS by 7.13\% to 12.78\%. Notably, unlike EquS~\cite{DBLP:journals/corr/abs-2511-09965}, LEADer leverages a state-aware trajectory pruning strategy to improve generation quality while achieving substantial sampling speedups, saving 3.04\% to 8.42\% of sampling time on average. These results demonstrate that our local epistemic uncertainty quantification directly strengthens the adaptive adjustment for complex local structures while significantly boosting inference efficiency.

\textbf{Qualitative Results.} To assess the restoration capability of LEADer under complex degradations, we compare the visual results of various baseline methods before and after its integration, as shown in Figure~\ref{Figure3}. We observe that baseline methods often suffer from detail loss or over-smoothing due to fixed prior weights. By integrating LEADer, the model can adaptively modulate the prior strength, which effectively suppresses artifacts while recovering sharper edges and more realistic textures, thus leading to a substantial improvement in visual fidelity.

To further illustrate the effect of the uncertainty-based active sampling strategy, we visualize the restoration trajectories of $\hat{\mathbf{x}}_{0|t}$ across different steps on the $4 \times$ super-resolution task, as shown in Figure~\ref{Figure4}. Due to its fixed step size, the baseline DDNM recovers details through a slow and uniform process, where both intermediate states and final outputs often exhibit high uncertainty. In contrast, LEADer perceives the current state and dynamically adjusts the sampling pace, which enables earlier convergence to clear local textures at lower steps. This enhanced sensitivity to detail not only improves data consistency but also allows the SATP strategy to safely skip steps, thereby reducing inference time.

\begin{table}[t]
    \centering
    \caption{Ablation study results of each component in LEADer.}
    \label{Table2}
    \setlength{\tabcolsep}{4pt}
    \renewcommand{\arraystretch}{1.0}
    \resizebox{0.98\linewidth}{!}{%
    \begin{tabular}{cccccccccccc}
        \toprule
        \multicolumn{3}{c}{DDNM}
          & \multicolumn{3}{c}{Bicub. SR $\times$ 4 $\left(\sigma_e = 0.05\right)$}
          & \multicolumn{3}{c}{Gaussian Deb. $\left(\sigma_e = 0.05\right)$}
          & \multicolumn{3}{c}{CS 25\% $\left(\sigma_e = 0.05\right)$} \\
        \cmidrule(lr){1-3}
        \cmidrule(lr){4-6}
        \cmidrule(lr){7-9}
        \cmidrule(lr){10-12}
        
        UCPM & SATP & T
          & {PSNR} $\uparrow$ & {LPIPS} $\downarrow$ & {Time} $\downarrow$
          & {PSNR} $\uparrow$ & {LPIPS} $\downarrow$ & {Time} $\downarrow$
          & {PSNR} $\uparrow$ & {LPIPS} $\downarrow$ & {Time} $\downarrow$ \\
        \midrule
        
        \xmark & \xmark & 100 & 29.19 & 0.082 & 3.87 s
                         & 30.60 & 0.072 & 3.90 s
                         & 26.14 & 0.124 & 3.83 s \\
                         
        \cmark & \xmark & 100 & 29.69 & 0.075 & 3.96 s
                         & 30.90 & 0.061 & 4.01 s
                         & 27.14. & 0.090 & 3.97 s \\
        \rowcolor{pink!10}
        \xmark & \xmark & 93 & 28.85 & 0.098 & \colorbox{pink!30} {3.67 s}
                         & 30.31 & 0.089 & \colorbox{pink!30} {3.83 s}
                         & 25.79 & 0.134 & \colorbox{pink!30} {3.75 s} \\         
        \rowcolor{pink!10}
        \xmark & \cmark & 100 & 29.17 & 0.084 & \colorbox{pink!30} {3.66 s}
                         & 30.58 & 0.077 & \colorbox{pink!30} {3.80 s}
                         & 26.10 & 0.125 & \colorbox{pink!30} {3.73 s} \\
        \rowcolor{bestrow}
        \cmark & \xmark & 93 & 29.35 & 0.090 & \colorbox{yellow!30} {3.65 s}
                         & 30.58 & 0.074 & \colorbox{yellow!30} {3.75 s}
                         & 26.74 & 0.106 & \colorbox{yellow!30} {3.70 s} \\
                         
        \rowcolor{bestrow}
        \cmark & \cmark & 100 & 29.69 & 0.077 & \colorbox{yellow!30} {3.63 s}
                         & 30.86 & 0.063 & \colorbox{yellow!30} {3.75 s}
                         & 27.06 & 0.094 & \colorbox{yellow!30} {3.69 s} \\
        \bottomrule
    \end{tabular}%
    }
    \vspace{-2mm}
\end{table}

\begin{table}[!t]
\centering
\caption{Hyperparameter sensitivity analysis on the information loss budget $B$.}
\footnotesize
\label{Table3}
\resizebox{0.48\textwidth}{!}{
\begin{tabular}{lcccccc}
\toprule
ImageNet 1K & \multicolumn{3}{c}{Bicub. SR $\times$ 4 $\left(\sigma_e = 0\right)$} & \multicolumn{3}{c}{Motion Deb. $\left(\sigma_e = 0.05\right)$} \\
\cmidrule(r){2-4} \cmidrule(l){5-7}
Setting & PSNR$\uparrow$ & LPIPS$\downarrow$ & Time$\downarrow$ & PSNR$\uparrow$ & LPIPS$\downarrow$ & Time$\downarrow$ \\
\midrule
$B=0$ & 25.41 & 0.143 & 8.56 & 25.31 & 0.196 & 9.83 \\
$B=0.001$ & 25.41 & 0.143 & 8.56 & 25.30 & 0.196 & 9.83 \\
$B=0.005$ & 25.38 & 0.144 & 8.45 & 25.26 & 0.199 & 9.74 \\
\rowcolor{blue!10} $B=0.01$  & 25.37 & 0.146 & 8.29 & 25.24 & 0.201 & 9.63 \\
$B=0.05$  & 24.83 & 0.161 & 4.54 & 24.71 & 0.226 & 5.21 \\
$B=0.1$   & 23.96 & 0.187 & 2.76 & 23.85 & 0.258 & 3.08 \\
\bottomrule
\end{tabular}
}
\vspace{-2mm}
\end{table}

\subsection{Ablation Study}
\textbf{Impact of Core Strategies.} To validate the effectiveness of the key strategies in the LEADer framework, we conduct ablation studies on three image restoration tasks using the DDNM~\cite{DBLP:conf/iclr/WangYZ23} baseline. As shown in Table~\ref{Table2}, introducing UCPM alone significantly improves the restoration quality, albeit with a slight increase in sampling time. Conversely, employing SATP alone effectively reduces inference time with virtually no quality degradation. When combined, they exhibit excellent complementarity, achieving superior restoration quality while simultaneously reducing the overall sampling time. This indicates that the effective local guidance provided by UCPM enhances the stability of reverse sampling, thereby enabling SATP to perform step-skipping more safely. Furthermore, to demonstrate the advantage of the SATP strategy over directly reducing sampling steps, we conduct a step ablation study. The results show that, under the condition of roughly equivalent sampling time, our active sampling strategy demonstrates significant robustness. These experimental results demonstrate that both UCPM and SATP are indispensable, yielding clear synergy and mutual enhancement across spatiotemporal dimensions.

\begin{figure}[!t]
\centering
\includegraphics[width=0.48\textwidth]{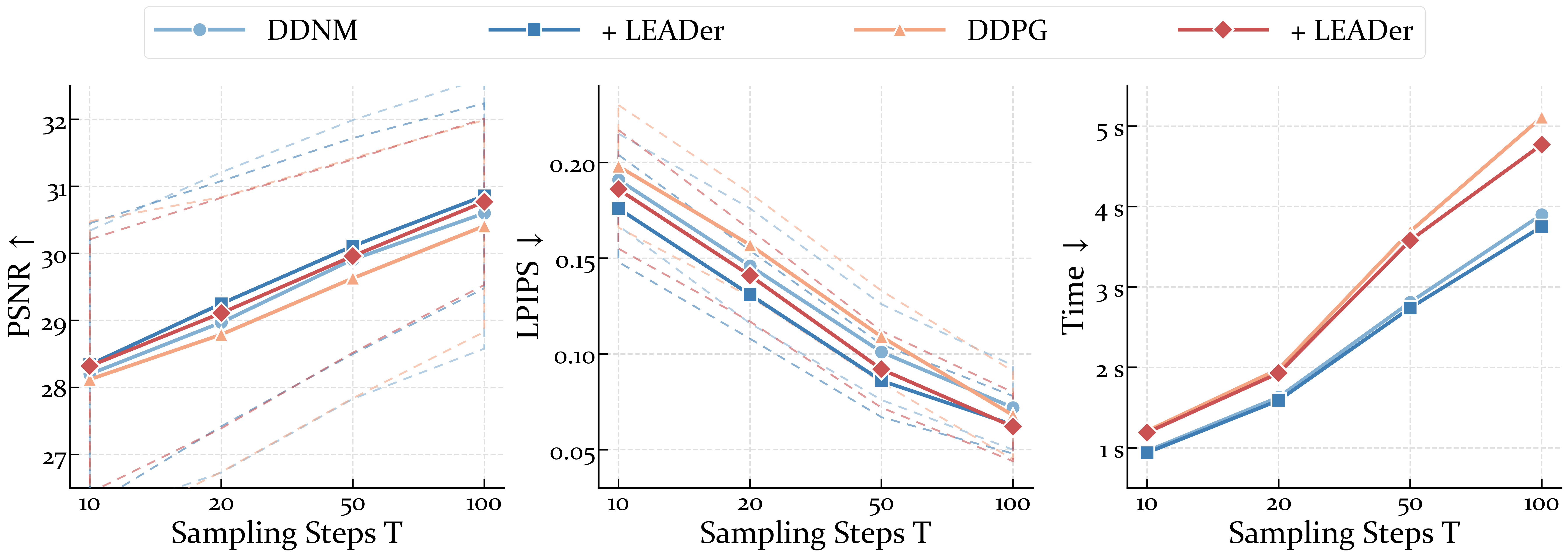}
\caption{Ablation study on different sampling steps. $\left( \text{Dataset: CelebA-HQ 1K, Task: Gaussion Deb. \& } \sigma = 0.05 \right)$}
\label{Figure5}
\vspace{-5mm}
\end{figure}

\noindent\textbf{Impact of Sampling Steps.} To evaluate the robustness of our method across varying sampling steps, we compare the performance of DDNM~\cite{DBLP:conf/iclr/WangYZ23} and DDPG~\cite{DBLP:conf/cvpr/GarberT24} with and without LEADer at $\text{T} \in \{10, 20, 50, 100\}$. As shown in Figure~\ref{Figure5}, introducing LEADer consistently improves image quality across all settings, thus demonstrating the effectiveness of our UCPM. In terms of efficiency, SATP consistently reduces inference time. However, under extremely low step counts, the runtime reduction becomes marginal. This is because the overall epistemic uncertainty remains excessively high under such sparse sampling; consequently, SATP adaptively performs fewer skip operations to preserve restoration quality. Overall, these results verify that LEADer exhibits strong robustness across different sampling steps.

\begin{table}[t]
    \centering
    \caption{Memory consumption comparison across five tasks.}
    \label{Table4}
    \setlength{\tabcolsep}{2pt}
    \renewcommand{\arraystretch}{1.00}
    \footnotesize
    \resizebox{0.45\textwidth}{!}{%
    \begin{tabular}{lccccc}
        \toprule
        Methods & Bicub. SR $\times$ 4 & Bicub. SR $\times$ 4 & Gaussian Deb. & Motion Deb. & CS 25\% \\
        & ($\sigma_e = 0$) & ($\sigma_e = 0.05$) & ($\sigma_e = 0.05$) & ($\sigma_e = 0.05$) & ($\sigma_e = 0.05$) \\
        \midrule
        DDNM~\cite{DBLP:conf/iclr/WangYZ23}
        & 2507 MB & 2513 MB & 2509 MB &  N / A & 2541 MB \\
        \rowcolor{green!15} + LEADer
        & 2509 MB & 2516 MB & 2515 MB & N / A & 2543 MB \\
        \addlinespace[2pt]
        \hdashline
        \addlinespace[2pt]
        DDPG~\cite{DBLP:conf/cvpr/GarberT24}
        & 2521 MB & 2524 MB & 2523 MB & 2959 MB & 2564 MB \\
        \rowcolor{green!15} + LEADer
        & 2526 MB & 2526 MB & 2530 MB & 2965 MB & 2567 MB \\
        \addlinespace[2pt]
        \hdashline
        \addlinespace[2pt]
        SITCOM~\cite{DBLP:conf/icml/AlkhouriLHD0RW25}
        & 15831 MB & 15863 MB & 15912 MB & 18478 MB & N / A \\
        \rowcolor{green!15} + LEADer
        & 15842 MB & 15872 MB & 15923 MB & 18485 MB & N / A \\
        \addlinespace[2pt]
        \hdashline
        \addlinespace[2pt]
        PIRP F$_{\mathrm{lb}}\text{G}_{-}$~\cite{DBLP:journals/kbs/YangZZZ26}
        & 2513 MB & 2520 MB & 2515 MB & 2948 MB & 2557 MB \\
        \rowcolor{green!15} + LEADer
        & 2516 MB & 2521 MB & 2521 MB & 2952 MB & 2561 MB \\
        \bottomrule
    \end{tabular}%
    }
    \vspace{-5mm}
\end{table}

\noindent\textbf{Impact of $B$.} The information loss budget $B$ in Equation~\eqref{eq18} governs the trade-off between restoration quality and computational efficiency. A larger $B$ tolerates higher local truncation errors, thereby encouraging SATP to execute more aggressive trajectory pruning. To investigate the impact of $B$, we conduct a sensitivity analysis using PIRP F$_{\text{lb}}\text{G}_{-}$~\cite{DBLP:journals/kbs/YangZZZ26} as the baseline on the ImageNet 1K dataset for the Bicubic SR $\times 4$ $\left(\sigma_e = 0\right)$ and Motion Deblurring $\left(\sigma_e = 0.05\right)$ tasks. As shown in Table~\ref{Table3}, a conservative budget $\left(B=0.001\right)$ yields quality improvements while restricting acceleration. Conversely, an overly aggressive budget $\left(B=0.1\right)$, despite significantly reducing inference time, leads to notable performance degradation. Based on these observations, we set $B=0.01$ as the default configuration for our SATP strategy to optimally balance restoration fidelity and sampling efficiency.

\subsection{Memory Consumption}
To evaluate the practical overhead of LEADer as a plug-and-play module, we compare the GPU memory consumption of various baseline methods before and after its integration, as shown in Table~\ref{Table4}. Evaluations across five tasks on the ImageNet 1K dataset reveal that despite introducing additional computations, LEADer incurs a marginal overall GPU memory increase of merely 0.07\% to 0.28\%. This indicates that LEADer can robustly enhance both the restoration quality and sampling efficiency of existing methods at an almost negligible memory cost.

\section{Conclusion}
\balance
In this paper, we identify a fundamental limitation of existing zero-shot DMIR methods, i.e., their reliance on fixed prior constraints and uniform sampling schedules, which often leads to local structural distortions and redundant computations. To address this, we propose LEADer, an active diffusion
sampling framework guided by local epistemic uncertainty. Spatially, we leverage pixel-level uncertainty to dynamically modulate prior strength, which effectively balances detail preservation and artifact suppression. Temporally, we quantify the stability of the generation process via uncertainty traces and adaptively prune trajectories within deterministic error bounds. Both theoretical analysis and extensive experiments demonstrate that LEADer can be seamlessly integrated as a plug-and-play module to consistently improve restoration quality and sampling efficiency, while preserving strict data consistency. In the future, it would be also interesting to investigate the application of this uncertainty-based active sampling mechanism to diffusion-model-based generative tasks.

\begin{acks}
This paper was supported by NSFC under Grant No. 62376108, 61402205, and Jiangsu University under Grant No. 13JDG085.
\end{acks}

\bibliographystyle{ACM-Reference-Format}
\balance
\bibliography{Reference}

@String{Computing = "Computing" }

@String{Computer = "{IEEE} Computer" }

@String{Springer = "Springer-Verlag" }

@inproceedings{DBLP:conf/icml/Sohl-DicksteinW15,
  author       = {Jascha Sohl{-}Dickstein and
                  Eric A. Weiss and
                  Niru Maheswaranathan and
                  Surya Ganguli},
  editor       = {Francis R. Bach and
                  David M. Blei},
  title        = {Deep Unsupervised Learning using Nonequilibrium Thermodynamics},
  booktitle    = {International Conference on Machine Learning},
  series       = {JMLR Workshop and Conference Proceedings},
  volume       = {37},
  pages        = {2256--2265},
  publisher    = {JMLR.org},
  year         = {2015},
}

@inproceedings{DBLP:conf/nips/HoJA20,
  author       = {Jonathan Ho and
                  Ajay Jain and
                  Pieter Abbeel},
  title        = {Denoising Diffusion Probabilistic Models},
  booktitle    = {Annual Conference
                  on Neural Information Processing Systems},
  year         = {2020},
}

@inproceedings{DBLP:conf/iclr/0011SKKEP21,
  author       = {Yang Song and
                  Jascha Sohl{-}Dickstein and
                  Diederik P. Kingma and
                  Abhishek Kumar and
                  Stefano Ermon and
                  Ben Poole},
  title        = {Score-Based Generative Modeling through Stochastic Differential Equations},
  booktitle    = {International Conference on Learning Representations},
  publisher    = {OpenReview.net},
  year         = {2021},
}

@inproceedings{DBLP:conf/nips/SongE19,
  author       = {Yang Song and
                  Stefano Ermon},
  title        = {Generative Modeling by Estimating Gradients of the Data Distribution},
  booktitle    = {Annual Conference
                  on Neural Information Processing Systems},
  pages        = {11895--11907},
  year         = {2019},

}

@inproceedings{DBLP:conf/cvpr/RombachBLEO22,
  author       = {Robin Rombach and
                  Andreas Blattmann and
                  Dominik Lorenz and
                  Patrick Esser and
                  Bj{\"{o}}rn Ommer},
  title        = {High-Resolution Image Synthesis with Latent Diffusion Models},
  booktitle    = {IEEE/CVF Conference on Computer Vision and Pattern Recognition},
  pages        = {10674--10685},
  publisher    = {IEEE},
  year         = {2022},
}

@inproceedings{DBLP:conf/nips/KawarEES22,
  author       = {Bahjat Kawar and
                  Michael Elad and
                  Stefano Ermon and
                  Jiaming Song},
  title        = {Denoising Diffusion Restoration Models},
  booktitle    = {Annual Conference on Neural Information Processing Systems},
  year         = {2022},
}

@inproceedings{DBLP:conf/cvpr/ZhuZLCWTG23,
  author       = {Yuanzhi Zhu and
                  Kai Zhang and
                  Jingyun Liang and
                  Jiezhang Cao and
                  Bihan Wen and
                  Radu Timofte and
                  Luc Van Gool},
  title        = {Denoising Diffusion Models for Plug-and-Play Image Restoration},
  booktitle    = {IEEE/CVF Conference on Computer Vision and Pattern Recognition},
  pages        = {1219--1229},
  publisher    = {IEEE},
  year         = {2023},
}

@inproceedings{DBLP:conf/iclr/WangYZ23,
  author       = {Yinhuai Wang and
                  Jiwen Yu and
                  Jian Zhang},
  title        = {Zero-Shot Image Restoration Using Denoising Diffusion Null-Space Model},
  booktitle    = {International Conference on Learning Representations},
  publisher    = {OpenReview.net},
  year         = {2023},
}

@inproceedings{DBLP:conf/cvpr/GarberT24,
  author       = {Tomer Garber and
                  Tom Tirer},
  title        = {Image Restoration by Denoising Diffusion Models with Iteratively Preconditioned Guidance},
  booktitle    = {IEEE/CVF Conference on Computer Vision and Pattern Recognition},
  pages        = {25245--25254},
  publisher    = {IEEE},
  year         = {2024},
}

@inproceedings{DBLP:conf/iclr/ChungKMKY23,
  author       = {Hyungjin Chung and
                  Jeongsol Kim and
                  Michael Thompson McCann and
                  Marc Louis Klasky and
                  Jong Chul Ye},
  title        = {Diffusion Posterior Sampling for General Noisy Inverse Problems},
  booktitle    = {International Conference on Learning Representations},
  publisher    = {OpenReview.net},
  year         = {2023},
}

@inproceedings{DBLP:conf/cvpr/ZhangIESW18,
  author       = {Richard Zhang and
                  Phillip Isola and
                  Alexei A. Efros and
                  Eli Shechtman and
                  Oliver Wang},
  title        = {The Unreasonable Effectiveness of Deep Features as a Perceptual Metric},
  booktitle={IEEE/CVF Conference on Computer Vision and Pattern Recognition},
  pages        = {586--595},
  year         = {2018},
}

@inproceedings{DBLP:conf/iccv/LiuLWT15,
  author       = {Ziwei Liu and
                  Ping Luo and
                  Xiaogang Wang and
                  Xiaoou Tang},
  title        = {Deep Learning Face Attributes in the Wild},
  booktitle    = {International Conference on Computer Vision},
  pages        = {3730--3738},
  publisher    = {IEEE Computer Society},
  year         = {2015},
}

@article{DBLP:journals/ijcv/RussakovskyDSKS15,
  author       = {Olga Russakovsky and
                  Jia Deng and
                  Hao Su and
                  Jonathan Krause and
                  Sanjeev Satheesh and
                  Sean Ma and
                  Zhiheng Huang and
                  Andrej Karpathy and
                  Aditya Khosla and
                  Michael S. Bernstein and
                  Alexander C. Berg and
                  Li Fei{-}Fei},
  title        = {ImageNet Large Scale Visual Recognition Challenge},
  journal      = {Int. J. Comput. Vis.},
  volume       = {115},
  pages        = {211--252},
  year         = {2015},
}

@inproceedings{DBLP:conf/iclr/SongME21,
  author       = {Jiaming Song and
                  Chenlin Meng and
                  Stefano Ermon},
  title        = {Denoising Diffusion Implicit Models},
  booktitle    = {International Conference on Learning Representations},
  publisher    = {OpenReview.net},
  year         = {2021},
}

@article{DBLP:journals/kbs/YangZZZ26,
  author       = {Yang Yang and
                  Xi Zhang and
                  Jiaqi Zhang and
                  Lanling Zeng},
  title        = {Parameterized image restoration with diffusion and gradient priors},
  journal      = {Knowledge Based Systems},
  volume       = {338},
  pages        = {115488},
  year         = {2026},
}

@inproceedings{DBLP:conf/icml/AlkhouriLHD0RW25,
  author       = {Ismail Alkhouri and
                  Shijun Liang and
                  Cheng{-}Han Huang and
                  Jimmy Dai and
                  Qing Qu and
                  Saiprasad Ravishankar and
                  Rongrong Wang},
  title        = {SITCOM: Step-wise Triple-Consistent Diffusion Sampling For Inverse Problems},
  booktitle    = {International Conference on Machine Learning},
  volume       = {267},
  publisher    = {PMLR / OpenReview.net},
  year         = {2025},
}

@inproceedings{DBLP:conf/nips/DhariwalN21,
  author       = {Prafulla Dhariwal and
                  Alexander Quinn Nichol},
  title        = {Diffusion Models Beat GANs on Image Synthesis},
  booktitle    = {Annual Conference            on Neural Information Processing Systems},
  pages        = {8780--8794},
  year         = {2021},
}

@inproceedings{DBLP:conf/cvpr/LugmayrDRYTG22,
  author       = {Andreas Lugmayr and
                  Martin Danelljan and
                  Andr{\'{e}}s Romero and
                  Fisher Yu and
                  Radu Timofte and
                  Luc Van Gool},
  title        = {RePaint: Inpainting using Denoising Diffusion Probabilistic Models},
  booktitle    = {IEEE/CVF Conference on Computer Vision and Pattern Recognition},
  pages        = {11451--11461},
  publisher    = {IEEE},
  year         = {2022},
}

@article{DBLP:journals/corr/abs-2511-09965,
  author       = {Chenxu Wu and
                  Qingpeng Kong and
                  Peiang Zhao and
                  Wendi Yang and
                  Wenxin Ma and
                  Fenghe Tang and
                  Zihang Jiang and
                  S. Kevin Zhou},
  title        = {Equivariant Sampling for Improving Diffusion Model-based Image Restoration},
  journal      = {CoRR},
  volume       = {abs/2511.09965},
  year         = {2025},
}

@article{DBLP:journals/pami/Wang0H21,
  author       = {Zhihao Wang and
                  Jian Chen and
                  Steven C. H. Hoi},
  title        = {Deep Learning for Image Super-Resolution: A Survey},
  journal      = {IEEE Transactions on Pattern Analysis and Machine Intelligence},
  volume       = {43},
  number       = {10},
  pages        = {3365--3387},
  year         = {2021},
}

@article{DBLP:journals/tmi/WangYMF18,
  author       = {Ge Wang and
                  Jong Chul Ye and
                  Klaus Mueller and
                  Jeffrey A. Fessler},
  title        = {Image Reconstruction is a New Frontier of Machine Learning},
  journal      = {IEEE Transactions on Medical Imaging},
  volume       = {37},
  number       = {6},
  pages        = {1289--1296},
  year         = {2018},
}

@article{DBLP:journals/tip/ZhangZCM017,
  author       = {Kai Zhang and
                  Wangmeng Zuo and
                  Yunjin Chen and
                  Deyu Meng and
                  Lei Zhang},
  title        = {Beyond a Gaussian Denoiser: Residual Learning of Deep CNN for Image Denoising},
  journal      = {IEEE Transactions on Image Processing},
  volume       = {26},
  number       = {7},
  pages        = {3142--3155},
  year         = {2017},
}

@inproceedings{DBLP:conf/iccvw/LiangCSZGT21,
  author       = {Jingyun Liang and
                  Jiezhang Cao and
                  Guolei Sun and
                  Kai Zhang and
                  Luc Van Gool and
                  Radu Timofte},
  title        = {SwinIR: Image Restoration Using Swin Transformer},
  booktitle    = {IEEE/CVF International Conference on Computer Vision Workshops},
  pages        = {1833--1844},
  publisher    = {IEEE},
  year         = {2021},

}

@inproceedings{DBLP:conf/cvpr/ZamirA0HK022,
  author       = {Syed Waqas Zamir and
                  Aditya Arora and
                  Salman Khan and
                  Munawar Hayat and
                  Fahad Shahbaz Khan and
                  Ming{-}Hsuan Yang},
  title        = {Restormer: Efficient Transformer for High-Resolution Image Restoration},
  booktitle    = {IEEE/CVF Conference on Computer Vision and Pattern Recognition},
  pages        = {5718--5729},
  publisher    = {IEEE},
  year         = {2022},
}

@inproceedings{DBLP:conf/eccv/ChenCZS22,
  author       = {Liangyu Chen and
                  Xiaojie Chu and
                  Xiangyu Zhang and
                  Jian Sun},
  title        = {Simple Baselines for Image Restoration},
  booktitle    = {European Conference on Computer Vision},
  series       = {Lecture Notes in Computer Science},
  volume       = {13667},
  pages        = {17--33},
  publisher    = {Springer},
  year         = {2022},
}

@article{DBLP:journals/pami/ZhangLZZGT22,
  author       = {Kai Zhang and
                  Yawei Li and
                  Wangmeng Zuo and
                  Lei Zhang and
                  Luc Van Gool and
                  Radu Timofte},
  title        = {Plug-and-Play Image Restoration With Deep Denoiser Prior},
  journal      = {IEEE Transactions on Pattern Analysis and Machine Intelligence},
  volume       = {44},
  number       = {10},
  pages        = {6360--6376},
  year         = {2022},
}

@inproceedings{DBLP:conf/iclr/0011S0E22,
  author       = {Yang Song and
                  Liyue Shen and
                  Lei Xing and
                  Stefano Ermon},
  title        = {Solving Inverse Problems in Medical Imaging with Score-Based Generative Models},
  booktitle    = {International Conference on Learning Representations},
  publisher    = {OpenReview.net},
  year         = {2022},
}

@inproceedings{DBLP:conf/nips/ChungSRY22,
  author       = {Hyungjin Chung and
                  Byeongsu Sim and
                  Dohoon Ryu and
                  Jong Chul Ye},
  title        = {Improving Diffusion Models for Inverse Problems using Manifold Constraints},
  booktitle    = {Annual Conference on Neural Information Processing Systems},
  year         = {2022},
}

@inproceedings{DBLP:conf/mm/WuHZ0LLZ0024,
  author       = {Hongjie Wu and
                  Linchao He and
                  Mingqin Zhang and
                  Dongdong Chen and
                  Kunming Luo and
                  Mengting Luo and
                  Jizhe Zhou and
                  Hu Chen and
                  Jiancheng Lv},
  title        = {Diffusion Posterior Proximal Sampling for Image Restoration},
  booktitle    = {ACM International Conference on Multimedia},
  pages        = {214--223},
  publisher    = {ACM},
  year         = {2024},
}

@inproceedings{DBLP:conf/mm/WuZHZ025,
  author       = {Hongjie Wu and
                  Mingqin Zhang and
                  Linchao He and
                  Ji{-}Zhe Zhou and
                  Jiancheng Lv},
  title        = {Enhancing Diffusion Model Stability for Image Restoration via Gradient
                  Management},
  booktitle    = {ACM International Conference on Multimedia},
  pages        = {10768--10777},
  publisher    = {ACM},
  year         = {2025},
}

@inproceedings{DBLP:conf/nips/ZhangZJLG24,
  author       = {Jiawei Zhang and
                  Jiaxin Zhuang and
                  Cheng Jin and
                  Gen Li and
                  Yuantao Gu},
  title        = {Unleashing the Denoising Capability of Diffusion Prior for Solving
                  Inverse Problems},
  booktitle    = {Annual Conference on Neural Information Processing Systems},
  year         = {2024},

}

@inproceedings{DBLP:conf/iclr/SongVMK23,
  author       = {Jiaming Song and
                  Arash Vahdat and
                  Morteza Mardani and
                  Jan Kautz},
  title        = {Pseudoinverse-Guided Diffusion Models for Inverse Problems},
  booktitle    = {International Conference on Learning Representations},
  publisher    = {OpenReview.net},
  year         = {2023},

}

@inproceedings{DBLP:conf/eccv/AlcalarA24,
  author       = {Yasar Utku Al{\c{c}}alar and
                  Mehmet Ak{\c{c}}akaya},
  title        = {Zero-Shot Adaptation for Approximate Posterior Sampling of Diffusion
                  Models in Inverse Problems},
  booktitle    = {European Conference on Computer Vision},
  series       = {Lecture Notes in Computer Science},
  volume       = {15141},
  pages        = {444--460},
  publisher    = {Springer},
  year         = {2024},
}

@article{li2024decoupled,
author       = {Xiang Li and
                  Soo Min Kwon and
                  Ismail R. Alkhouri and
                  Saiprasad Ravishankar and
                  Qing Qu},
  title        = {Decoupled Data Consistency with Diffusion Purification for Image Restoration},
  journal      = {CoRR},
  volume       = {abs/2403.06054},
  year         = {2024},
}

@inproceedings{DBLP:conf/cvpr/ZhangCBMA025,
  author       = {Bingliang Zhang and
                  Wenda Chu and
                  Julius Berner and
                  Chenlin Meng and
                  Anima Anandkumar and
                  Yang Song},
  title        = {Improving Diffusion Inverse Problem Solving with Decoupled Noise Annealing},
  booktitle    = {IEEE/CVF Conference on Computer Vision and Pattern Recognition},
  pages        = {20895--20905},
  publisher    = {IEEE},
  year         = {2025},
}

@inproceedings{DBLP:conf/nips/KarrasAAL22,
  author       = {Tero Karras and
                  Miika Aittala and
                  Timo Aila and
                  Samuli Laine},
  title        = {Elucidating the Design Space of Diffusion-Based Generative Models},
  booktitle    = {Annual Conference on Neural Information Processing Systems},
  year         = {2022},
}

@inproceedings{DBLP:conf/nips/0011ZB0L022,
  author       = {Cheng Lu and
                  Yuhao Zhou and
                  Fan Bao and
                  Jianfei Chen and
                  Chongxuan Li and
                  Jun Zhu},
  title        = {DPM-Solver: {A} Fast {ODE} Solver for Diffusion Probabilistic Model
                  Sampling in Around 10 Steps},
  booktitle    = {Annual Conference on Neural Information Processing Systems},
  year         = {2022},
}

@article{DBLP:journals/spm/McCannJU17,
  author       = {Michael T. McCann and
                  Kyong Hwan Jin and
                  Michael Unser},
  title        = {Convolutional Neural Networks for Inverse Problems in Imaging: A Review},
  journal      = {IEEE Signal Processing Magazine},
  volume       = {34},
  number       = {6},
  pages        = {85--95},
  year         = {2017},
}

@article{DBLP:journals/pami/DongLHT16,
  author       = {Chao Dong and
                  Chen Change Loy and
                  Kaiming He and
                  Xiaoou Tang},
  title        = {Image Super-Resolution Using Deep Convolutional Networks},
  journal      = {IEEE Transactions on Pattern Analysis and Machine Intelligence},
  volume       = {38},
  number       = {2},
  pages        = {295--307},
  year         = {2016},
}

@article{DBLP:journals/pami/CroitoruHIS23,
  author       = {Florinel{-}Alin Croitoru and
                  Vlad Hondru and
                  Radu Tudor Ionescu and
                  Mubarak Shah},
  title        = {Diffusion Models in Vision: {A} Survey},
  journal      = {IEEE Transactions on Pattern Analysis and Machine Intelligence},
  volume       = {45},
  number       = {9},
  pages        = {10850--10869},
  year         = {2023},
}

@article{DBLP:journals/csur/YangZSHXZZCY24,
  author       = {Ling Yang and
                  Zhilong Zhang and
                  Yang Song and
                  Shenda Hong and
                  Runsheng Xu and
                  Yue Zhao and
                  Wentao Zhang and
                  Bin Cui and
                  Ming{-}Hsuan Yang},
  title        = {Diffusion Models: A Comprehensive Survey of Methods and Applications},
  journal      = {ACM Computing Surveys},
  volume       = {56},
  number       = {4},
  pages        = {105:1--105:39},
  year         = {2024},
}

@article{efron2011tweedie,
  title={Tweedie’s formula and selection bias},
  author={Efron, Bradley},
  journal={Journal of the American Statistical Association},
  volume={106},
  number={496},
  pages={1602--1614},
  year={2011},
}

@article{stein1981estimation,
  title={Estimation of the mean of a multivariate normal distribution},
  author={Stein, Charles M},
  journal={The Annals of Statistics},
  pages={1135--1151},
  year={1981},
  publisher={JSTOR}
}

@inproceedings{DBLP:conf/cvpr/BlauM18,
  author       = {Yochai Blau and
                  Tomer Michaeli},
  title        = {The Perception-Distortion Tradeoff},
  booktitle    = {IEEE Conference on Computer Vision and Pattern Recognition},
  pages        = {6228--6237},
  publisher    = {Computer Vision Foundation / IEEE Computer Society},
  year         = {2018},

}

@inproceedings{DBLP:conf/icml/NingSPCC23,
  author       = {Mang Ning and
                  Enver Sangineto and
                  Angelo Porrello and
                  Simone Calderara and
                  Rita Cucchiara},
  title        = {Input Perturbation Reduces Exposure Bias in Diffusion Models},
  booktitle    = {International Conference on Machine Learning},
  series       = {Proceedings of Machine Learning Research},
  pages        = {26245--26265},
  publisher    = {PMLR},
  year         = {2023},
}

@article{he2023fast,
  title={Fast and stable diffusion inverse solver with history gradient update},
  author={He, Linchao and Yan, Hongyu and Luo, Mengting and Wu, Hongjie and Luo, Kunming and Wang, Wang and Du, Wenchao and Chen, Hu and Yang, Hongyu and Zhang, Yi and others},
  journal      = {CoRR},
  volume       = {abs/2307.12070},
  year={2023}
}

@inproceedings{DBLP:conf/cvpr/FeiLPZYLZ023,
  author       = {Ben Fei and
                  Zhaoyang Lyu and
                  Liang Pan and
                  Junzhe Zhang and
                  Weidong Yang and
                  Tianyue Luo and
                  Bo Zhang and
                  Bo Dai},
  title        = {Generative Diffusion Prior for Unified Image Restoration and Enhancement},
  booktitle    = {IEEE/CVF Conference on Computer Vision and Pattern Recognition},
  pages        = {9935--9946},
  publisher    = {IEEE},
  year         = {2023},

}

@inproceedings{DBLP:conf/iclr/MardaniSKV24,
  author       = {Morteza Mardani and
                  Jiaming Song and
                  Jan Kautz and
                  Arash Vahdat},
  title        = {A Variational Perspective on Solving Inverse Problems with Diffusion
                  Models},
  booktitle    = {International Conference on Learning Representations},
  publisher    = {OpenReview.net},
  year         = {2024},
}

@inproceedings{DBLP:conf/nips/ChungKY23,
  author       = {Hyungjin Chung and
                  Jeongsol Kim and
                  Jong Chul Ye},
  title        = {Direct Diffusion Bridge using Data Consistency for Inverse Problems},
  booktitle    = {Annual Conference on Neural Information Processing Systems},
  year         = {2023},

}

@article{DBLP:journals/tip/DabovFKE07,
  author       = {Kostadin Dabov and
                  Alessandro Foi and
                  Vladimir Katkovnik and
                  Karen O. Egiazarian},
  title        = {Image Denoising by Sparse 3-D Transform-Domain Collaborative Filtering},
  journal      = {IEEE Transactions on Image Processing},
  volume       = {16},
  number       = {8},
  pages        = {2080--2095},
  year         = {2007},
}

@inproceedings{DBLP:conf/cvpr/GuZZF14,
  author       = {Shuhang Gu and
                  Lei Zhang and
                  Wangmeng Zuo and
                  Xiangchu Feng},
  title        = {Weighted Nuclear Norm Minimization with Application to Image Denoising},
  booktitle    = {IEEE Conference on Computer Vision and Pattern Recognition},
  pages        = {2862--2869},
  publisher    = {IEEE Computer Society},
  year         = {2014},
}

@inproceedings{DBLP:conf/cvpr/LedigTHCCAATTWS17,
  author       = {Christian Ledig and
                  Lucas Theis and
                  Ferenc Huszar and
                  Jose Caballero and
                  Andrew Cunningham and
                  Alejandro Acosta and
                  Andrew P. Aitken and
                  Alykhan Tejani and
                  Johannes Totz and
                  Zehan Wang and
                  Wenzhe Shi},
  title        = {Photo-Realistic Single Image Super-Resolution Using a Generative Adversarial
                  Network},
  booktitle    = {IEEE Conference on Computer Vision and Pattern Recognition},
  pages        = {105--114},
  publisher    = {IEEE Computer Society},
  year         = {2017},
}

@article{DBLP:journals/corr/abs-2411-15295,
  author       = {Darshan Thaker and
                  Abhishek Goyal and
                  Ren{\'{e}} Vidal},
  title        = {Frequency-Guided Posterior Sampling for Diffusion-Based Image Restoration},
  journal      = {CoRR},
  volume       = {abs/2411.15295},
  year         = {2024},
}

@inproceedings{DBLP:conf/cvpr/0011GF00KW25,
  author       = {Chong Wang and
                  Lanqing Guo and
                  Zixuan Fu and
                  Siyuan Yang and
                  Hao Cheng and
                  Alex C. Kot and
                  Bihan Wen},
  title        = {Reconciling Stochastic and Deterministic Strategies for Zero-shot
                  Image Restoration using Diffusion Model in Dual},
  booktitle    = {IEEE/CVF Conference on Computer Vision and Pattern Recognition},
  pages        = {23207--23216},
  publisher    = {Computer Vision Foundation / IEEE},
  year         = {2025},

}

@article{DPG,
  author       = {Yi Wang and
                  Lanling Zeng and
                  Jiaqi Zhang and
                  Yang Yang},
  title        = {Zero-Shot Diffusive Image Restoration With Consistency},
  journal      = {Signal Processing},
  volume       = {248},
  pages        = {110705},
  year         = {2026},
}

@article{lan2025mappo,
  title={MaPPO: Maximum a Posteriori Preference Optimization with Prior Knowledge},
  author={Lan, Guangchen and Zhang, Sipeng and Wang, Tianle and Zhang, Yuwei and Zhang, Daoan and Wei, Xinpeng and Pan, Xiaoman and Zhang, Hongming and Han, Dong-Jun and Brinton, Christopher G},
  journal={arXiv preprint arXiv:2507.21183},
  year={2025}
}

@article{zheng2022leveraging, 
title={Leveraging local and global cues for visual tracking via parallel interaction network}, 
author={Zheng, Yaozong and Zhong, Bineng and Liang, Qihua and Tang, Zhenjun and Ji, Rongrong and Li, Xianxian}, 
journal={IEEE Transactions on Circuits and Systems for Video Technology}, 
volume={33}, 
number={4}, 
pages={1671--1683}, 
year={2022}, 
publisher={IEEE} 
}

@article{he2026cinematte,
  title={Cinematte: Background matting for virtual production and beyond},
  author={He, Yuanjian and Zhang, Chen and Chen, Fasheng and Cao, Jiangbo},
  journal={arXiv preprint arXiv:2605.18328},
  year={2026}
}

@article{AgentTraces,
  author       = {Yiqi Wang and
                  Jiaqi Zhang and
                  Taotao Cai and
                  Zirui Liu and
                  Qingqiang Sun and
                  Zequn Sun and
                  Zhangkai Wu and
                  Manqing Dong and
                  Mingkai Zheng and
                  Xuefei Yin and
                  Yanming Zhu},
  title        = {From Agent Traces to Trust: {A} Survey of Evidence Tracing and Execution
                  Provenance in {LLM} Agents},
  journal      = {CoRR},
  volume       = {abs/2606.04990},
  year         = {2026},
}

@article{DG-SCL,
  author={Bingbing Gu and 
          Saihua Cai and 
          Jing Wang and 
          Zhuole Li and 
          Xiheng Jia and 
          Jiaqi Zhang},
  title={DG-SCL: Diffusion-Guided Semantic Contrastive Learning for Imbalanced Malicious Traffic Detection},
  journal={Information Sciences},
  pages={123851},
  year={2026},
}

@inproceedings{tao2024cntools,
title = {Cntools: A Computational Toolbox for Cellular Neighborhood Analysis from Multiplexed Images},
author = {Tao, Yicheng and Feng, Fan and Luo, Xin and Reihsmann, Conrad V. and Hopkirk, Alexander L. and Cartailler, Jean-Philippe and Brissova, Marcela and Parker, Stephen C. J. and Saunders, Diane C. and Liu, Jie},
booktitle = {PLoS Computational Biology},
volume = {20},
number = {8},
pages = {e1012344},
year = {2024},
publisher = {Public Library of Science}
}

@ARTICLE{11622841, 
author={Zhang, Fan and Fan, Shiming and Wang, Hua}, journal={IEEE Transactions on Knowledge and Data Engineering}, 
title={Rethinking Activation Function: A Simple Path to Efficient and Accurate Time Series Forecasting}, 
year={2026}, 
}

@inproceedings{zhang2026if,
  title={What If We Let Forecasting Forget? A Sparse Bottleneck for Cross-Variable Dependencies},
  author={Zhang, Fan and Fan, Shiming and Wang, Hua},
  booktitle={Forty-third International Conference on Machine Learning},
  year={2026}
}

@article{zhang2026drive,
  title={L-Drive: Beyond a Single Mapping-Latent Context Drives Time Series Forecasting},
  author={Zhang, Fan and Chen, Shijun and Wang, Hua},
  journal={arXiv preprint arXiv:2605.17730},
  year={2026}
}

\clearpage

\appendix

\section{More Experiments}

\begin{figure}[!th]
\centering
\includegraphics[width=0.48\textwidth]{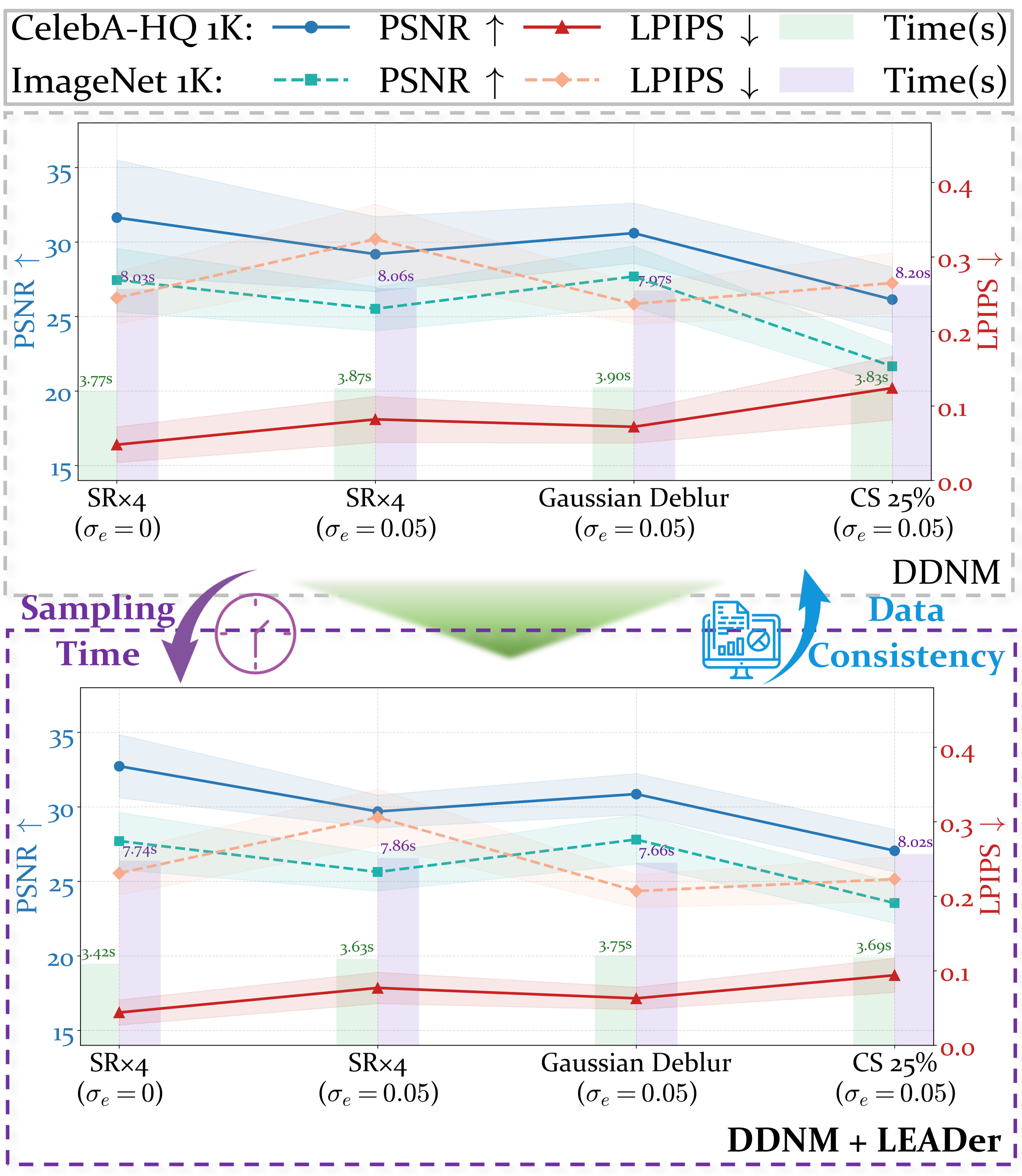}
\caption{Comparison of PSNR, LPIPS, and sampling time between DDNM and DDNM+LEADer on CelebA-HQ 1K and ImageNet 1K across different restoration tasks.}
\label{Figure6}
\end{figure}

We further compare DDNM and DDNM+LEADer in terms of PSNR, LPIPS, and sampling time on CelebA-HQ 1K and ImageNet 1K, as shown in Figure~\ref{Figure6}. The results further demonstrate that LEADer improves restoration quality while reducing sampling overhead.

\begin{figure}[!b]
\centering
\includegraphics[width=0.48\textwidth]{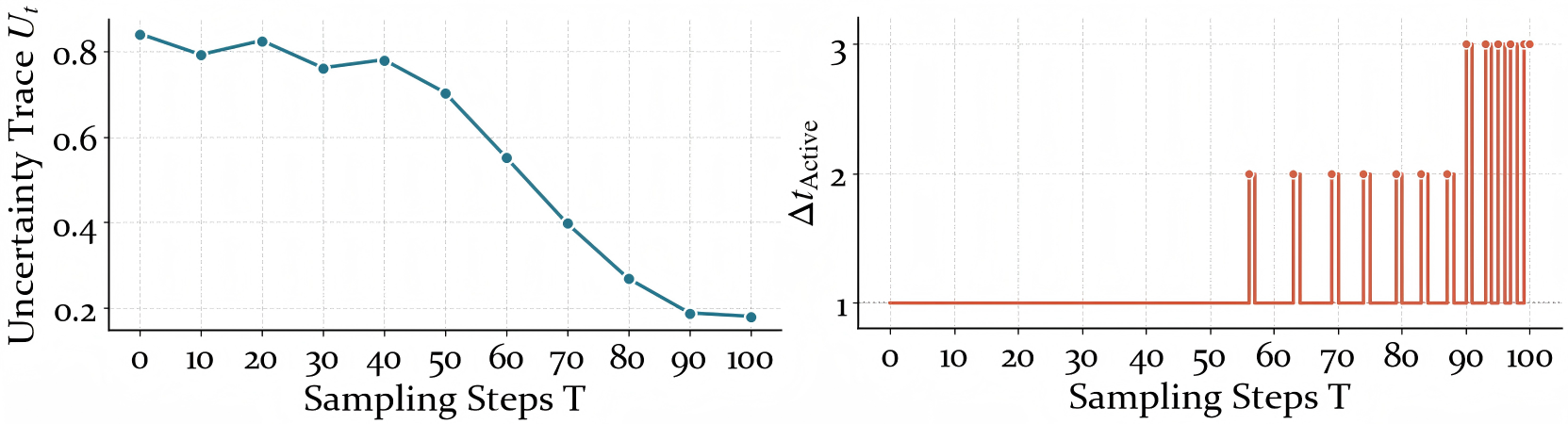}
\caption{Time-varying \(U_t\) and \(\Delta t_{\text{Active}}\) during SATP.}
\label{Figure7}
\end{figure}

We visualize the evolution of $U_t$ and $\Delta t_{\mathrm{Active}}$ throughout the sampling process, as shown in Figure~\ref{Figure7}. At the early sampling stage, $U_t$ remains high and exhibits pronounced fluctuations. SATP therefore does not perform additional step skipping, avoiding error accumulation during this unstable stage. As sampling proceeds, $U_t$ gradually decreases and stabilizes, while $\Delta t_{\mathrm{Active}}$ adaptively increases accordingly. These results show that SATP does not simply reduce the number of uniform sampling steps. Instead, it dynamically allocates the sampling budget according to the current uncertainty and skips redundant iterations once the process becomes stable, providing more direct evidence for its adaptive acceleration mechanism.

We further report the average FID across five restoration tasks on both datasets. As shown in Table~\ref{Table_FID}, LEADer reduces the Avg.~FID on CelebA-HQ 1K from 46.39 to 25.64 and on ImageNet 1K from 65.28 to 47.37. This result is consistent with the improvement in LPIPS, further indicating that LEADer effectively enhances the perceptual realism of restored images.

\section{Proof}
\subsection{Proof of Proposition 4.1}
\begin{proof}
Given the noiseless physical observation model $\mathbf{y} = \mathbf{A}\mathbf{x}_0$, it necessarily holds that $\mathbf{y} \in \text{Range}(\mathbf{A})$. 
By the properties of the Moore-Penrose pseudoinverse, the matrix $\mathbf{A}\mathbf{A}^{\dagger}$ is the orthogonal projection matrix onto the range of the operator $\mathbf{A}$, denoted as $\text{Range}(\mathbf{A})$. Thus, for any $\mathbf{y} \in \text{Range}(\mathbf{A})$, we have:
$$\mathbf{A}\mathbf{A}^{\dagger}\mathbf{y} = \mathbf{y}$$

Meanwhile, based on the definition of the null-space projection matrix $P_N = \mathbf{I} - \mathbf{A}^{\dagger}\mathbf{A}$ and the pseudoinverse identity $\mathbf{A}\mathbf{A}^{\dagger}\mathbf{A} = \mathbf{A}$, left-multiplying by the degradation operator $\mathbf{A}$ yields:
$$\mathbf{A}P_N = \mathbf{A}(\mathbf{I} - \mathbf{A}^{\dagger}\mathbf{A}) = \mathbf{A} - \mathbf{A}\mathbf{A}^{\dagger}\mathbf{A} = \mathbf{A} - \mathbf{A} = \mathbf{0}$$

Substituting these operator properties into the modulated estimation formulation of UCPM, $\hat{\mathbf{x}} = \mathbf{A}^{\dagger}\mathbf{y} + P_N \mathbf{x}_{Prior}$, and left-multiplying both sides by $\mathbf{A}$, we obtain:
$$\begin{aligned} \mathbf{A}\hat{\mathbf{x}} &= \mathbf{A} \left( \mathbf{A}^{\dagger}\mathbf{y} + P_N \mathbf{x}_{Prior} \right) \\ &= \mathbf{A}\mathbf{A}^{\dagger}\mathbf{y} + \mathbf{A}P_N \mathbf{x}_{Prior} \\ &= \mathbf{y} + \mathbf{0} \cdot \mathbf{x}_{Prior} \\ &= \mathbf{y} \end{aligned}$$

This demonstrates that the prior modulation variable $\mathbf{x}_{Prior}$, induced by epistemic uncertainty, is constrained within the null space of the operator $\mathbf{A}$ by $P_N$. Consequently, it does not interfere with the existing physical observations. Thus, the data fidelity residual is:
$$||\mathbf{y} - \mathbf{A}\hat{\mathbf{x}}||_2^2 = ||\mathbf{y} - \mathbf{y}||_2^2 = 0$$

This completes the proof.
\end{proof}

\begin{table}[!b]
\centering
\caption{Average FID comparison on CelebA-HQ 1K and ImageNet 1K.}
\label{Table_FID}
\setlength{\tabcolsep}{2pt}
\renewcommand{\arraystretch}{1.00}
\footnotesize
\resizebox{0.48\textwidth}{!}{%
\begin{tabular}{lcc}
\toprule
Method & CelebA-HQ 1K Avg. FID & ImageNet 1K Avg. FID \\
\midrule
DDNM
& 46.39 & 65.28 \\
\rowcolor{green!15} + LEADer
& 25.64 & 47.37 \\
\bottomrule
\end{tabular}%
}
\end{table}

\subsection{Proof of Proposition 4.2}
\begin{proof}
Consider the probability flow ODE for the reverse diffusion process: $d\mathbf{x}/dt = f(\mathbf{x}, t)$. By assumption, the drift field $f$ satisfies $L$-Lipschitz continuity in the state space:
$$||f(\mathbf{x}_1, t) - f(\mathbf{x}_2, t)||_2 \le L ||\mathbf{x}_1 - \mathbf{x}_2||_2, \quad \forall \mathbf{x}_1, \mathbf{x}_2$$

Let $t_k$ and $t_{k+1}$ be adjacent time steps, with a step size of $\Delta t_k = t_{k+1} - t_k$. The single-step exact evolution of the true solution $\mathbf{x}(t)$ and the single-step update of the SATP numerical solution $\tilde{\mathbf{x}}$ satisfy, respectively:  
$$\mathbf{x}(t_{k+1}) = \mathbf{x}(t_k) + \int_{t_k}^{t_{k+1}} f(\mathbf{x}(s), s) ds$$
$$\tilde{\mathbf{x}}_{k+1} = \tilde{\mathbf{x}}_k + \Delta t_k f(\tilde{\mathbf{x}}_k, t_k)$$

Let $e_k = ||\mathbf{x}(t_k) - \tilde{\mathbf{x}}_k||_2$ denote the global evolution error at time $t_k$. Following numerical ODE theory, we define the local truncation error at the $k$-th step, $\boldsymbol{\tau}_k$, as:
$$\boldsymbol{\tau}_k = \mathbf{x}(t_{k+1}) - \left[ \mathbf{x}(t_k) + \Delta t_k f(\mathbf{x}(t_k), t_k) \right]$$

The SATP mechanism strictly bounds this local truncation error such that $||\boldsymbol{\tau}_k||_2 \le C B \Delta t_k$. Combining the definition of $\boldsymbol{\tau}_k$ with the triangle inequality, we can expand and bound $e_{k+1}$ as follows:
$$\begin{aligned} e_{k+1} &= ||\mathbf{x}(t_{k+1}) - \tilde{\mathbf{x}}_{k+1}||_2 \\ &= ||\mathbf{x}(t_k) + \Delta t_k f(\mathbf{x}(t_k), t_k) + \boldsymbol{\tau}_k - \tilde{\mathbf{x}}_k - \Delta t_k f(\tilde{\mathbf{x}}_k, t_k)||_2 \\ &\le ||\mathbf{x}(t_k) - \tilde{\mathbf{x}}_k||_2 + \Delta t_k ||f(\mathbf{x}(t_k), t_k) - f(\tilde{\mathbf{x}}_k, t_k)||_2 + ||\boldsymbol{\tau}_k||_2 \end{aligned}$$

Applying the Lipschitz continuity condition and the truncation error bound yields:
$$e_{k+1} \le e_k + L \Delta t_k e_k + C B \Delta t_k = (1 + L \Delta t_k) e_k + C B \Delta t_k$$
Using the inequality $1 + x \le e^x$, this discrete error recurrence relation can be further bounded by:
$$e_{k+1} \le e^{L \Delta t_k} e_k + C B \Delta t_k$$

Assuming perfect sampling of the initial prior state ($e_0 = 0$), we can unroll the recurrence relation from $k=0$ up to the final time $t_N$:
$$\begin{aligned} e_N &\le \sum_{j=0}^{N-1} e^{L \sum_{i=j+1}^{N-1} \Delta t_i} C B \Delta t_j \\ &= C B \sum_{j=0}^{N-1} e^{L(t_N - t_{j+1})} \Delta t_j \end{aligned}$$

Since the integrand $e^{L(t_N - s)}$ is monotonically decreasing on $s \in [t_j, t_{j+1}]$, the right-hand side can be strictly upper-bounded by the corresponding definite integral:
$$e_N \le C B \int_0^{t_N} e^{L(t_N - s)} ds$$

Evaluating this integral analytically gives:
$$e_N \le \frac{C B}{L} \left[ -e^{L(t_N - s)} \right]_0^{t_N} = \frac{C B}{L} (e^{L t_N} - 1)$$

Given that the total generation time is $t_N \le T$, and the exponential function is monotonically increasing, we have $e_N \le \frac{C B}{L}(e^{LT} - 1)$. Meanwhile, note that the terminal evolution error $e_N$ is exactly the $\ell_2$-norm deviation between the continuous exact solution $\mathbf{x}^{Dense}$ and the SATP discrete solution $\mathbf{x}^{LEADer}$, i.e., $e_N = ||\mathbf{x}^{Dense} - \mathbf{x}^{LEADer}||_2$. 

Substituting this into the above inequality, we obtain the final global error bound:
$$||\mathbf{x}^{Dense} - \mathbf{x}^{LEADer}||_2 \le \frac{C \cdot B}{L}(e^{LT} - 1)$$
This completes the proof.
\end{proof}

\end{document}